\documentclass[10pt,journal,compsoc]{IEEEtran}

\ifCLASSOPTIONcompsoc
  \usepackage[nocompress]{cite}
\else
  \usepackage{cite}
\fi

\usepackage{array}
\usepackage{amssymb}
\usepackage{amsmath}
\usepackage{booktabs}
\usepackage{color}

\def\revision#1{{\color{black}{#1}}}
\def\newpara#1{\textbf{{#1}}}

\ifCLASSINFOpdf
  \usepackage[pdftex]{graphicx}
  \graphicspath{{../pdf/}{../jpeg/}}
  \DeclareGraphicsExtensions{.pdf,.jpeg,.png}
\else
\fi

\begin{document}
%
\title{StyleTalk++: A Unified Framework for Controlling the Speaking Styles of Talking Heads}

\author{Suzhen Wang\IEEEauthorrefmark{1}, Yifeng Ma\IEEEauthorrefmark{1}, Yu Ding\IEEEauthorrefmark{2}, Zhipeng Hu, Changjie Fan,  Tangjie Lv, Zhidong Deng, Xin Yu

\IEEEcompsocitemizethanks{
\IEEEcompsocthanksitem{\IEEEauthorrefmark{1}these authors contributed equally to this work.}
\IEEEcompsocthanksitem{\IEEEauthorrefmark{2}corresponding author.}

\IEEEcompsocthanksitem Suzhen Wang, Yu Ding, Tanjie Lv, Changjie Fan and Zhipeng Hu are with Fuxi AI Lab, Netease, Hangzhou, Zhejiang, China. E-mail: \{wangsuzhen, dingyu01, hzlvtangjie, fanchangjie, zphu\}@corp.netease.com.
\IEEEcompsocthanksitem Yifeng Ma and Zhidong Deng are with Department of Computer Science and Technology, BNRist, THUAI, State Key Laboratory of Intelligent Technology and Systems, Tsinghua University, China. E-mail: \{mayf18@mails., michael@\}tsinghua.edu.cn.
\IEEEcompsocthanksitem Xin Yu, is with the School of Computer Science, the University of Queensland, Brisbane, Australia. E-mail: xin.yu@uq.edu.au
}
\thanks{Preliminary versions of this work were published in IJCAI 2021~\cite{wang2021audio2head}, AAAI 2022~\cite{wang2022one} and AAAI 2023~\cite{ma2023styletalk}.}
}

%
%

\markboth{Journal of \LaTeX\ Class Files,~Vol.~XX, No.~X, XX XXXX}%
{Shell \MakeLowercase{\textit{et al.}}: StyleTalk++: A Unified Framework For Controlling the Speaking Styles of Talking Heads}
%

\IEEEtitleabstractindextext{%
\begin{abstract}

Individuals have unique facial expression and head pose styles that reflect their personalized speaking styles. Existing one-shot talking head methods cannot capture such personalized characteristics and therefore fail to produce diverse speaking styles in the final videos. To address this challenge, we propose a one-shot style-controllable talking face generation method that can obtain speaking styles from reference speaking videos and drive the one-shot portrait to speak with the reference speaking styles and another piece of audio. Our method aims to synthesize the style-controllable coefficients of a 3D Morphable Model (3DMM), including facial expressions and head movements, in a unified framework. Specifically, the proposed framework first leverages a style encoder to extract the desired speaking styles from the reference videos and transform them into style codes. Then, the framework uses a style-aware decoder to synthesize the coefficients of 3DMM from the audio input and style codes. During decoding, our framework adopts a two-branch architecture, which generates the stylized facial expression coefficients and stylized head movement coefficients, respectively. After obtaining the coefficients of 3DMM, an image renderer renders the expression coefficients into a specific person's talking-head video. Extensive experiments demonstrate that our method generates visually authentic talking head videos with diverse speaking styles from only one portrait image and an audio clip.

\end{abstract}

}

\maketitle

\IEEEdisplaynontitleabstractindextext

%
\IEEEpeerreviewmaketitle

\ifCLASSOPTIONcompsoc
\IEEEraisesectionheading{\section{Introduction}\label{sec:introduction}}
\else
\section{Introduction}
\label{sec:introduction}
\fi

%
%
%
%

\IEEEPARstart{A}{udio-driven} photo-realistic talking head generation has drawn growing attention due to its broad applications in virtual human creation, visual dubbing, and short video creation.  The past few years have witnessed tremendous progress in accurate lip synchronization~\cite{prajwal2020lip, wang2022one}, head pose generation~\cite{zhou2021pose, wang2021audio2head} and high-fidelity video generation~\cite{zhang2021flow, yin2022styleheat}. However, existing \emph{one-shot} based works pay less attention to modeling diverse speaking styles, thus failing to produce expressive talking head videos with various styles. 

The speaking styles of individuals consist of both facial expression style and head pose style, which respectively represent the spatial and temporal co-activations of full facial expressions and head poses. In real-world scenarios, different individuals may speak the same utterance with significantly diverse personalized speaking styles. Due to such significant diversities, creating controllable talking heads that showcase specific styles remains a great challenge, particularly in one-shot settings. Previous works~\cite{wang2020mead,sinha2022emotion} have denoted speaking style simply as discrete emotion classes, which is insufficient for representing flexible speaking styles. Even though recent methods~\cite{ji2022eamm,liang2022expressive} can control upper facial expressions by incorporating an additional emotional source video, they only transfer upper facial motion characteristics at a 
 static frame level, ignoring the temporal dynamics of speaking styles. Therefore, a universal spatio-temporal representation of speaking styles is highly desirable.

 \begin{figure*}[ht]
\centering
\includegraphics[width=0.98\textwidth]{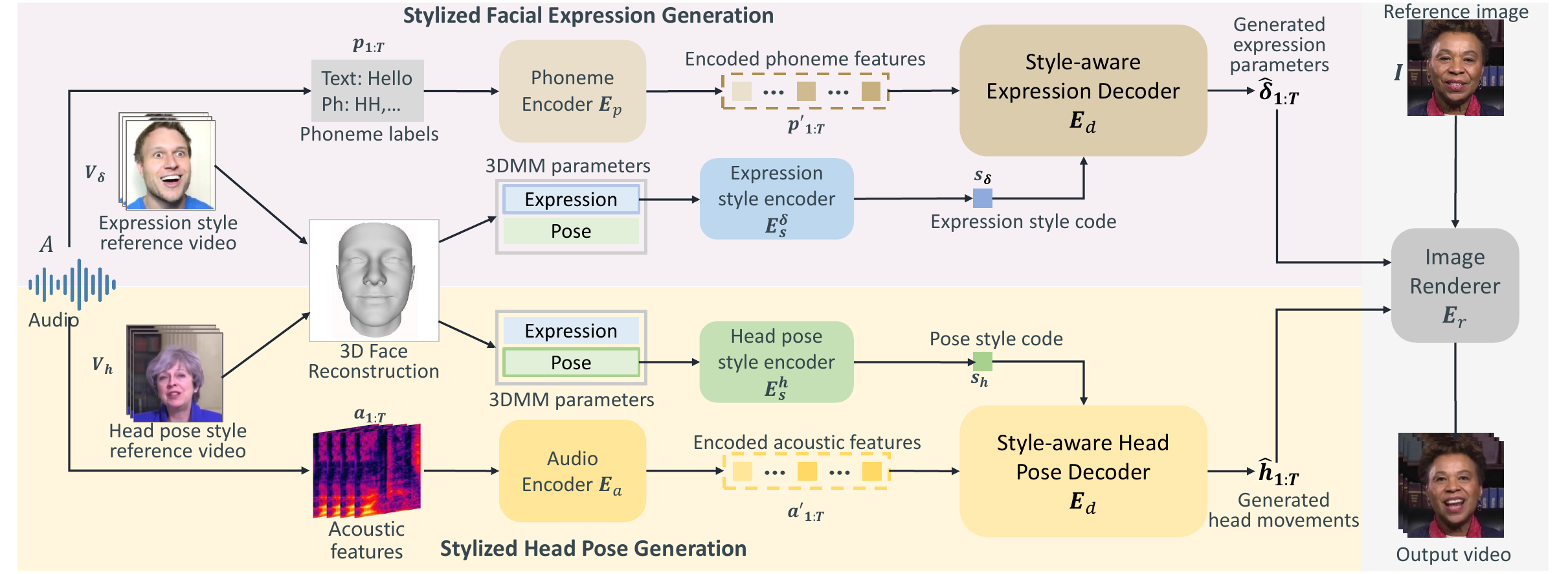}
\caption{ Illustration of StyleTalk++. Our method can control both facial expression and head pose styles in the generated talking faces using a unified style-controllable framework. These styles can be reflected in two additional style reference videos, including the expression style and head pose style videos, which can be \textbf{the same}. 
The unified style-controllable framework is extended into two branches: (1) The stylized expression generation branch first extracts sequential 3DMM expression parameters from the expression style reference video $\boldsymbol{V}_{\delta}$ using the 3D face reconstruction module, and then feeds them into the expression style encoder $\boldsymbol{E}^{\delta}_s$ to obtain the expression style code ${\boldsymbol{s}_{\delta}}$. A phoneme encoder $\boldsymbol{E}_p$ encodes phoneme labels into phoneme features $\boldsymbol{p}'_{1,T}$. Then, the style-aware expression decoder $\boldsymbol{E}_d$ generates the stylized expression parameters ${\hat{\boldsymbol{\delta}}_{1:T}}$ with $\boldsymbol{s_{\delta}}$ and $\boldsymbol{p}'_{1:T}$. (2) Similarly, the stylized head pose generation branch first extracts sequential head poses from the head pose style reference video and obtains the head pose style code ${\boldsymbol{s}_{h}}$. An acoustic encoder $\boldsymbol{E}_a$ encodes acoustic features into latent features $\boldsymbol{a}'_{1,T}$. Then, we use a style-aware head pose decoder $\boldsymbol{E}_d$ to generate the stylized head movements ${\hat{\boldsymbol{h}}_{1:T}}$ from $\boldsymbol{s_{h}}$ and $\boldsymbol{a}'_{1:T}$. Finally, the image renderer $\boldsymbol{E}_r$ takes the assembled ${\hat{\boldsymbol{\delta}}_{1:T}}$ and ${\hat{\boldsymbol{h}}_{1:T}}$, and the identity reference image $\boldsymbol{I}^r$ as input, and generates the output video.
}
\label{fig:pipeline}
\end{figure*} 

In this paper, we propose a new method called \textbf{StyleTalk++} that can learn a comprehensive representation of speaking style from a talking video. Our approach aims to create stylized and realistic talking videos for a one-shot speaker image, where the speaker delivers the specified audio content with the extracted speaking style from style reference videos. To achieve this, our method utilizes a unified, style-controllable framework that first extracts the speaking style from a reference video and then embeds it into the audio-driven generated coefficients of a 3D Morphable Model (3DMM). These coefficients include facial expression and head pose parameters, and as a result, the unified framework is instantiated into two branches for stylized facial expression generation and head pose generation, respectively. Finally, an image renderer takes facial animations, head poses, and the reference image as inputs to generate photorealistic talking faces.

First, we design a universal style encoder to model the motion patterns of facial expressions and head poses in arbitrary reference style videos. The purpose of the style encoder is to extract the latent style codes (i.e., expression style code or head pose style code) from the sequential 3DMM expression or head pose of the reference style videos. To achieve this, the style encoder utilizes a transformer encoder to study the spatio-temporal co-activation patterns of the input sequential parameters. It then employs a self-attention pooling layer \cite{safari2020self} to embed these patterns into the style codes.   We also introduce a triplet constraint on the style code space, which allows the universal style encoder to be applied to unseen style clips. Additionally, we observe that the learned style codes lie in a semantically meaningful space.

Afterwards, the unified framework introduces a style-aware decoder that synthesizes stylized animation parameters from audio, based on the style codes.  To better incorporate the style codes into the generated animation parameters, the style-aware decoder employs the transformer decoder as the backbone and uses the style code as the query. \revision{Leveraging cross attention, the style code can guide the model to focus on closely associating the audio representations with a specific style, thereby enhancing the synthesis of stylized animations.} Facial expressions and head movements show distinct motion characteristics, prompting us to develop different decoding strategies. For stylized facial expressions, we propose adaptively generating kernel weights of the feed-forward layers in the transformer decoder conditioned on the style code. This improves lip-sync in various styles and yields more convincing facial expressions. In the stylized head pose decoder, we introduce recurrence into the transformer and predict head movements step-by-step. This allows for the creation of a natural head motion sequence that matches the audio rhythm in different styles.


In summary, our proposed method, StyleTalk++, presents an innovative approach to creating stylized talking videos. Our unified, style-controllable framework enables the extraction and embedding of speaking styles from style reference videos, resulting in the production of natural and photorealistic talking faces in various speaking styles. Extensive experiments demonstrate that our method can generate photorealistic talking faces with diverse speaking styles while satisfying accurate lip synchronization, convincing facial expressions, and natural head movement. We believe that our approach provides a significant contribution to the fields of expressive talking face generation and stylized animation generation.

\section{Related Work}

\subsection{Audio-Driven Talking Head Generation}
With the increasing demand for virtual human creation, driving talking heads with audio~\cite{zhu2021deep,chen2020comprises} has attracted considerable attention. Audio-driven methods can be classified into two categories: person-specific and person-agnostic methods. 

\subsubsection{Person-specific Methods}
Person-specific methods \cite{suwajanakorn2017synthesizing,fried2019text,thies2020neural,li2021write,zhang20213d,song2020everybody,ji2021audio,lah2021lipsync3d,zhang2021facial,guo2021ad,liu2022semantic} are only applicable to speakers seen during training. \cite{thies2020neural} produces a high-quality talking video of a target person by using a latent 3D face model. \cite{li2021write} propose a novel text-based talking-head video generation framework that synthesizes high-fidelity facial expressions and head motions in accordance with contextual sentiments, speech rhythm, and pauses. \cite{lah2021lipsync3d} proposes to convert arbitrary talking head video footage into a normalized space that decouples 3D pose, geometry, texture, and lighting, thereby enabling data-efficient learning and versatile high-quality lip-sync synthesis for video and 3D applications. \cite{zhang2021facial} proposes FACIAL-GAN to jointly learn explicit (facial expression) and implicit (head poses, eye blinks) attributes from audio features. Recently, \cite{guo2021ad} and \cite{liu2022semantic} introduced neural radiance fields for high-fidelity talking head generation.

\subsubsection{Person-agnostic Methods}
Person-agnostic methods aim to generate talking head videos in a one-shot setting.
Early methods \cite{chung2017you,song2018talking,chen2018lip,song2018talking,vougioukas2019realistic,das2020speech,zhou2019talking,chen2019hierarchical} focus only on creating accurate mouth movements that are synchronized with the speech content.
\cite{zhou2019talking} learns a joint audio-visual representation through audio-visual speech discrimination by associating several supervisions. \cite{chen2019hierarchical} proposes a novel cascade network structure to reduce the effects of the sound-irrelevant visual dynamics in the image space and explicitly constructs high-level representation from the audio signal and guides video generation using the inferred representation.

With the development of deep learning, a number of methods \cite{wiles2018x2face,chen2020talking,zhou2020makelttalk,prajwal2020lip,zhang2021flow,wang2021audio2head,zhou2021pose,wang2022one} have been developed to produce more natural talking faces by taking facial expressions and head poses into consideration. \cite{chen2020talking} proposes a 3D-aware generative network to explicitly model head motion and facial expressions. \cite{zhang2021flow} first simultaneously produces movements of the mouth, eyebrows, and head pose, and then transforms the animation into videos using a flow-guided video generator. \cite{zhou2021pose} implicitly learns pose information directly from reference videos without using intermediate representations. \cite{prajwal2020lip} only generates the lip-synced mouth region while using the head pose directly from the original videos. \cite{wang2021audio2head} exploits a keypoint-based dense motion field representation to produce natural head motions while keeping non-face regions stable. 
\cite{wang2022one} extracts consistent audio-visual correlations from a specific speaker and achieves satisfactory visual quality and accurate lip-sync. However, although the aforementioned methods can generate videos for arbitrary speakers, none of these methods can create expressive talking head videos.

\subsection{Expressive Talking Head Generation}
Although expressive facial expressions are crucial in vivid talking head generation, only a few methods\cite{sadoughi2019speech,wang2020mead,wu2021imitating,ji2021audio,sinha2022emotion,ji2022eamm,liang2022expressive，gururani2023space} take it into consideration.
\cite{wang2020mead} build emotional talking head dataset MEAD and propose an emotional talking head generation baseline. 
\cite{ji2021audio} extract disentangled content and emotional information from audio, and then produce videos guided by the predicted landmarks. However, determining emotions only from audio  may lead to ambiguities\cite{ji2022eamm}, limiting the applicability of an emotional talking face model.
\cite{wang2020mead}, \cite{sinha2022emotion} and \cite{gururani2023space} create emotion-controllable talking faces by employing explicit emotion labels as input, which drop the formulation of personalized differences in speaking styles.
\cite{ji2022eamm} and \cite{liang2022expressive} generate expressive talking heads by transferring the expressions in an additional emotional source video to the target speaker frame-by-frame. 
To sum up, none of the previous works captures the spatial and temporal co-activations of facial expressions.

\subsection{Audio-driven Head Movement Generation}
Talking head videos with natural head movements appear more realistic. Numerous prior works have explored the generation of audio-driven head movements. One category of methods, such as those presented in \cite{zhang20213d,zhang2021facial,lu2021live,yi2020audio}, only model the audio-to-head-movement mapping reflected in a target speaker video. However, these methods require re-training or finetuning when applied to unseen speakers. Another category of methods, presented in \cite{chen2020talking,zhou2020makelttalk,zhang2021flow,wang2021audio2head}, learn the audio-to-head-movement mapping from videos of various speakers and can be applied to unseen speakers without re-training. However, it should be noted that \cite{zhou2020makelttalk} can only produce head movements that slightly swing around the initial pose in the reference image. While \cite{zhang2021flow} and \cite{wang2021audio2head} can generate natural head movements from audio, they all have limitations in producing head movements with diverse styles, ignoring the diversity of head pose styles.

\subsection{Extension to Our Prior Work}
This paper builds upon our prior research, which includes Audio2Head~\cite{wang2021audio2head}, AVCT~\cite{wang2022one}, and StyleTalk~\cite{ma2023styletalk}. Our current research extends the idea of controlling StyleTalk in a unified framework that can control both expression and head pose styles. Note that our previous works Audio2Head and AVCT can generate natural head poses and accurate lip-sync, and StyleTalk focuses on generating diverse facial expressions. In this work, we extend this research to explore how to produce accurate lip-sync and natural head poses across diverse styles. Additionally, we adopt the batched sequential training paradigm proposed in AVCT to achieve realistic talking-head generation.  Furthermore, we conduct more comprehensive experiments to validate the effectiveness of our improvements.



\section{METHODOLOGY}

In this paper, we propose StyleTalk++ for generating the style-controllable talking faces with four inputs: (1) the reference image  $\boldsymbol{I}^r$ of the target speaker; (2)  the audio clip $\boldsymbol{A}$ of length $T$ providing the speech content; (3) the expression style reference talking video $\boldsymbol{V}_{\delta} = \boldsymbol{I}^{\delta}_{1:N}$ of length $N$, referred to as the expression style clip; (4) the head pose style reference talking video $\boldsymbol{V}_h = \boldsymbol{I}^h_{1:M}$ of length $M$, referred to as the pose style clip. The expression style clip and pose style clip may be the same. Our method can create photo-realistic taking videos $\boldsymbol{Y} = \hat{\boldsymbol{I}}_{1:T}$ in which the target speaker speaks the speech content with the facial expression style reflected in the expression style clip and head pose style reflected in the head pose style clip. 


To generate style-controllable talking faces, we begin by generating the stylized coefficients of a 3DMM, which are then rendered into videos of a specific speaker. We propose a unified style-controllable framework for generating stylized facial expressions and head poses. The framework extracts the speaking style from style reference videos and embeds it into the audio-driven generated coefficients. As shown in Figure \ref{fig:pipeline}, this framework has been extended into two branches for stylized facial expression generation and stylized head pose generation, respectively. Each branch comprises a 3D face reconstruction module, a style encoder, an audio encoder, and a style-controllable decoder. Note that the two branches share the same 3D face reconstruction module, and their style encoder and audio encoder adopt similar network architectures. Therefore, we first introduce the 3D face reconstruction module in Section~\ref{face_renco} and the universal style encoder in Section~\ref{style_encoder}. We then describe the process of generating stylized head poses and facial expressions in Sections~\ref{style_head_pose_gen} and \ref{style_expression_gen}, respectively. Finally, an image renderer is used to convert the generated 3DMM coefficients and reference image into a video. We describe this renderer in Section~\ref{renderer}.

\subsection{3D Face Reconstruction}\label{face_renco}
With a 3DMM \cite{deng2019accurate}, the face shape $\mathbf{S}$ can be represented by an affine model:
\begin{equation}
    \mathbf{S} = \mathbf{S}(\boldsymbol{\delta},\boldsymbol{\phi})=
    \Bar{\mathbf{S}} + \mathbf{B}_{exp}\boldsymbol{\delta}+\mathbf{B}_{id}\boldsymbol{\phi},
\end{equation}
where $\Bar{\mathbf{S}}$ is the average face shape, $\mathbf{B}_{id}$ and $\mathbf{B}bb_{exp}$ are the PCA bases of identity and expression respectively; $\boldsymbol{\delta} \in \mathbb{R}^{64}$ and $\boldsymbol{\phi} \in \mathbb{R}^{80}$ are the corresponding coefficient vectors for a specific 3D face. We adopt the popular 2009 Basel Face Model~\cite{paysan20093d} for $\Bar{\mathbf{S}}$ and $\mathbf{B}_{id}$, and use the expression bases $\mathbf{B}_{exp}$ of \cite{guo2018cnn}. In addition, the head rotation and translation are expressed as $\mathcal{R} \in \mathbb{R}^{3}$ and  $\mathcal{T} \in \mathbb{R}^{3}$. 

An off-the-shelf 3D face reconstruction model~\cite{deng2019accurate} is employed for extracting the 3DMM coefficients from portrait images.
We employ a subset of 3DMM expression parameters 
$\boldsymbol{\delta}_{1:N}$ as the facial representation. Given the style clips $\boldsymbol{V}_{\delta}$ and $\boldsymbol{V}_h$, the 3D face reconstruction module extracts the sequential facial expression parameters $\boldsymbol{\delta}_{1:N}$ and head poses $\boldsymbol{h}_{1:M}$, where ${\boldsymbol{h}_i = \{\mathcal{R}_i,\mathcal{T}_i\}}$.






\subsection{Universal Style Encoder}\label{style_encoder}
Previous methods for synthesizing stylized facial animations and head pose only transfer the static motions of the static images \cite{ji2022eamm, liang2022expressive}. Unlike these methods, our approach aims to model the dynamic motion patterns that can guide synthesis. We develop a universal style encoder $\mathbf{E}_s$ to extract the spatio-temporal speaking style reflected in the style clip. The speaking styles in the corresponding sequential facial expression parameters $\boldsymbol{\delta}_{1:N}$ and head poses $\boldsymbol{h}_{1:M}$ are represented as expression style code $\boldsymbol{s}_{\delta}$ and head pose style code $\boldsymbol{s}_{h}$.  We use a transformer encoder that takes the sequential $\boldsymbol{\delta}_{1:N}$ or $\boldsymbol{h}_{1:M}$ as input tokens. The encoder models the temporal correlation between tokens and outputs the style vectors of each token, $\boldsymbol{s}'_{1:N}$. Since the speaking style in a video clip can be identified by a few typical frames, we employ a self-attention pooling layer \cite{safari2020self} to aggregate the style information over the style vectors. Specifically, this layer uses an additive attention-based mechanism that computes the token-level attention weights using a feed-forward network. The token-level attention weights represent the frame-level impact on the video-level style code. By summing all the style vectors multiplied by their attention weights, we obtain the final style code $\boldsymbol{s} \in \mathbb{R}^{d_s}$:
\begin{equation}
    \boldsymbol{s} = \operatorname{softmax}(W_s H)H^T,
\end{equation}
where $W_s\in \mathbb{R}^{1 \times d_s}$ is a trainable parameter, $H = [\boldsymbol{s}_1,...\boldsymbol{s}_N] \in \mathbb{R}^{d_s \times N}$ is the sequence of encoded features, $d_s$ is the dimension of each style vector. Using the same approach, we get the expression style code $\boldsymbol{s}_{\delta}$ and head poses style code $\boldsymbol{s}_{h}$.

Our intuition is that the style codes corresponding to similar speaking styles should cluster in the style space. To achieve this, we utilize a triplet constraint on the style codes generated by the style encoder. To apply this constraint, we begin by randomly sampling two additional style clips, $\boldsymbol{V}_c^p$ and $\boldsymbol{V}_c^n$, which reflect similar and dissimilar speaking styles, respectively, to a given style clip $\boldsymbol{V}_c$. Corresponding style codes $\boldsymbol{s}_c$, $\boldsymbol{s}_c^p$, and $\boldsymbol{s}_c^n$ are then extracted from the triplet-paired videos $\{\boldsymbol{V}_c, \boldsymbol{V}_c^p, \boldsymbol{V}_c^n\}$. Finally, we enforce the constraint on their distances in the style space using the triplet loss \cite{dong2018triplet}:
\begin{equation}
\mathcal{L}_{trip} = \max \{{\|\boldsymbol{s}_c-  \boldsymbol{s}_c^p \|}_2 - {\|\boldsymbol{s}_c-  \boldsymbol{s}_c^n \|}_2 + \gamma, 0\},
\end{equation}
where $\gamma$ is the margin parameter and is set to 5. Note that $\boldsymbol{s}_c$ may represent either head pose or facial expression style code. 

\begin{figure}\centering
 \includegraphics[width=0.45\textwidth]{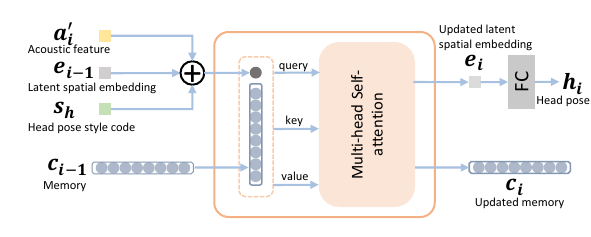}
 \caption{Illustration of the $i$-th step of the style-aware head pose decoder. The latent spatial embedding $\boldsymbol{e}_i$ and memory $\boldsymbol{c}_i$ are the intermediate features in this decoder and will be updated at each step.}
 \label{Fig: head_motion}
\end{figure}

\subsection{Stylized Head Pose Generation}\label{style_head_pose_gen}
\subsubsection{Acoustic Encoder}
In the stylized head pose generation branch, the audio is processed into acoustic features, which provide essential rhythm and intonation information related to head movement.
As preprocessing, the input raw audio $\mathbf{A}$ is first converted to an acoustic feature sequence $\boldsymbol{a}_{1:T}$. Each $\boldsymbol{a}_i$ refers to an acoustic feature frame. To match the video frequency, which is sampled at 25 frames per second, each $\boldsymbol{a}_i \in \mathbb{R}^{164}$ is composed of acoustic features extracted from four successive sliding windows. A total of 41 acoustic features are extracted from each sliding window, including 13 Mel Frequency Cepstrum Coefficients (MFCCs), 26 Mel-filterbank energy features (FBANK), pitch, and voicelessness. The sliding window has a window size of 25ms and a step size of 10ms. We utilize another transformer encoder as our acoustic encoder, denoted as $\mathbf{E}_a$, to extract acoustic embeddings $\boldsymbol{a}'_i \in \mathbb{R}^{256}$ from $\boldsymbol{a}_{1:T}$.

\subsubsection{Style-Aware Head pose Decoder}
The style-aware head pose decoder $\mathbf{E}^h_{d}$ generates stylized head pose movements by taking sequential acoustic embeddings $\boldsymbol{a}'_{1:T}$ as input and conditioning on the head pose style code $\boldsymbol{s}_{h}$ in a sequence-to-sequence manner. However, real-life head movements are non-deterministic and rely not only on the long-term audio rhythm but also on the immediate head pose state. To fit this, we develop our $\mathbf{E}^h_{d}$ based on Transformer-XL~\cite{dai2019transformer}, which introduces recurrence into the Transformer by employing hidden states from previous segments as memory for the current segment. By using Transformer-XL, we establish temporal correlations between head movements and audio features, enabling us to produce a head motion sequence that matches the audio rhythm naturally.

We recurrently predict the head movements step by step using $\mathbf{E}^h_{d}$, as shown in Figure~\ref{Fig: head_motion}. At each time step $i$, we begin by appending the head pose style code $\boldsymbol{s}_{h}$ with an absolute position embedding~\cite{vaswani2017attention}. Next, we concatenate the style code with the audio feature $\boldsymbol{a}'_i$ and feed them into the Transformer-XL. The Transformer-XL combines the input features with memory along the length dimension to calculate the multi-head self-attention. It then outputs the spatial embedding $e_i$, which encodes the current head pose's spatial state, and updates the memory simultaneously. 

To account for the local motion state more accurately, we introduce a spatial embedding transition by attaching the previous output spatial embedding $e_{i-1}$ to the integrated input features. This technique enables the decoder to produce more stable head movements and better synchronization with audio. This procedure is formulated as:
\begin{equation}
    (\boldsymbol{c}_i,\boldsymbol{e}_i) = \mathbf{TransXL}(\boldsymbol{c}_{i-1},\boldsymbol{a}'_i \oplus \boldsymbol{e}_{i-1} \oplus \boldsymbol{s}_h),
\end{equation}
where $c_i$ is the memory of step $i$ in Transformer-XL, $\oplus$ means concatenation, $\mathbf{TransXL}$ means Transformer-XL. 

Finally, we use a fully-connected (FC) layer to decode $e_i$ to head pose $\boldsymbol{h}_i \in \mathbb{R}^6$, where 3 dimensions are for rotation and 3 for translation. Our head motion predictor can handle an arbitrary length of audio input. To ensure better alignment of the generated poses with the camera space of the reference speaker image, the decoder takes an extra initial pose $\boldsymbol{h}_r$ as input. This initial pose can be extracted from the reference speaker image using the 3D face reconstruction model, or it can be a specified head pose. To map $\boldsymbol{h}_r$ to the initial spatial embedding $\boldsymbol{e}_0$, we utilize a four-layer feed-forward network.
By conditioning on different head pose style codes, our method can create diverse natural-looking head movements while matching the audio rhythm. The stylized head pose generation process can be formulated as :
\begin{equation}
    \boldsymbol{\hat{h}}_{1:T} = \mathbf{E}^h_d(a'_{1:T}, \boldsymbol{s}_h, \boldsymbol{h}_r)
\end{equation}
where ${\boldsymbol{\hat{h}}}_{1:T}$ means the generated head pose sequence.

\subsubsection{Head Pose Objective Function Design}
This section outlines our training approach for the style-aware head pose generation module. Specifically, we jointly train the acoustic encoder $\mathbf{E}_a$, the head pose style encoder $\mathbf{E}^h_{s}$, and the style-aware head pose decoder $\mathbf{E}^h_d$ using the following loss function design:

\newpara{Head Pose Reconstruction Constraint.} As the mapping from audio to head motion is one-to-many, the commonly used L1 and L2 losses are unsuitable for supervising the reconstruction of the head pose. Instead, we use SSIM loss as the reconstruction loss $\mathcal{L}^h_{rec}$. SSIM (Structural Similarity Index Measure)~\cite{wang2004image} is a well-known image quality metric that evaluates similarity between two images based on their luminance, contrast, and structure. To be specific, we consider the head motion sequence $\boldsymbol{h}_{1:T} \in \mathbb{R}^{6 \times T}$ as an image of size $6 \times T$, and use SSIM loss to impose structural constraints on it, thereby preserving the velocity, frequency, and amplitude of the head motion. SSIM loss is formulated as:
\begin{equation}
    \mathcal{L}_{SSIM} =  1 - \frac{(2 \mu \hat{\mu} + C_1) (2 {cov}+C_2))}{(\mu^{2} + \hat{\mu}^{2} + C_1)(\sigma^{2} + \hat{\sigma}^{2} + C_2))}.
\end{equation}
where $\hat{\mu}$ and $\hat{\sigma}$ are the mean and standard deviation of the generated head pose sequence $\hat{\boldsymbol{h}}_{1:T}$, and ${\mu}$ and ${\sigma}$ are that of the ground truth head pose sequence. $cov$ is the covariance. $C_1$ and $C_2$ are two small constants. To train our model to reconstruct the head pose, we utilize SSIM loss as the reconstruction loss $\mathcal{L}^h_{rec}$.

\newpara{Head Pose Temporal Discriminator.}
In order to improve the smoothness of the generated head pose sequence ${\boldsymbol{\hat{h}}}_{1:T}$, we use a head pose temporal discriminator $\mathbf{D}^h_{tem}$, which learns to differentiate between real and fake input head pose sequence. Specifically, we modify the 2D PatchGAN discriminator~\cite{goodfellow2014generative,isola2017image,zhu2017unpaired,yu2017face,yu2017hallucinating,yu2018face}, which is designed to process patches of the input image rather than the entire image, into a 1D window discriminator that focuses on the temporal window of the input sequence. Following the same network structure as the vanilla PatchGAN discriminator, our $\mathbf{D}^h_{tem}$ performs 1D convolution instead of 2D convolution on the input head pose sequence along the temporal axis. This helps to classify whether 70 × 70 overlapping head pose windows are real or fake and in turn, improves the smoothness of the generated head pose sequence. Additionally, we employ LSGAN~\cite{mao2017least} to calculate the adversarial loss:
\begin{equation}
    \mathcal{L}^h_{tem} = \|\mathbf{D}^h_{tem}({\boldsymbol{\hat{h}}}_{1:T})-1\|_2 .
\end{equation}

\newpara{Head Pose Style Discriminator. }
To ensure consistency in the style between the generated head movements and the specified head pose style, we propose a head pose style discriminator $\mathbf{D}^h_{style}$, which shares a similar network architecture to $\mathbf{D}^h_{tem}$. The primary objective of $\mathbf{D}^h_{style}$ is to distinguish whether the input head poses belong to the specified head pose style. $\mathbf{D}^h_{style}$ takes the integrated head pose sequence and style code as input. Specifically, the style code is first repeated $T$ times and appended with position embeddings, and then concatenated with the head pose sequence along the time dimension. During training, $\mathbf{D}^h_{style}$ learns to minimize the following objective:
\begin{equation}
\begin{split}
    \mathcal{L}^{h,D}_{style} = \| \mathbf{D}^h_{style}(\boldsymbol{h}_{1:T}, \boldsymbol{s}_h) -1 \|_2 +\\
    + \| \mathbf{D}^h_{style}(\boldsymbol{h}_{1:T}, \boldsymbol{s}^n_h) -0 \|_2,
\end{split}
\end{equation}
where $\boldsymbol{s}_h$ denote the head pose style reflected in the ground truth head pose $\boldsymbol{h}_{1:T}$, $\boldsymbol{s}^n_h$ denotes another head pose style that is not similar to $\boldsymbol{s}_h$. \revision{When training the stylized head pose generation module, the same speech input is used along with different style codes. To ensure that the style of the generated head movements aligns with the specified head pose style, we utilize a style adversarial loss.} The style adversarial loss is defined as:
\begin{equation}
    \begin{split}
        \mathcal{L}^h_{style} = \| \mathbf{D}^h_{style}(\mathbf{E}^h_d(a'_{1:T}, \boldsymbol{s}_h, \boldsymbol{h}_r), \boldsymbol{s}_h) -1 \|_2 +\\
    + \| \mathbf{D}^h_{style}(\mathbf{E}^h_d(a'_{1:T}, \boldsymbol{s}^n_h, \boldsymbol{h}_r), \boldsymbol{s}^n_h) -1 \|_2.
    \end{split}
\end{equation}
\revision{Furthermore, because all style reference lengths are randomly sampled during the training process, these lengths may differ from that of the generated head pose, which is determined by the length of the input audio. This indicates that the generated head pose cannot be a mere replication of the reference when style adversarial loss ensures that the generated head movements conform to the specified head pose style.}

\newpara{Full Objective.} Our full objective for the stylized head pose generation module is:
\begin{equation}
    \begin{split}
        \mathcal{L}^h_{total}=\lambda^h_{rec} \mathcal{L}^h_{rec}+\lambda^h_{trip} \mathcal{L}^h_{trip} +\\
    + \lambda^h_{tem} \mathcal{L}^h_{tem}+\lambda^h_{style} \mathcal{L}^h_{style},
    \end{split}
\end{equation}
where $\mathcal{L}^h_{trip}$ is the triplet loss for the head pose style code, as introduced in Section~\ref{style_encoder}, and the hyper-parameters$\lambda^h_{rec}$, $\lambda^h_{trip}$, $\lambda^h_{tem}$ and $\lambda^h_{style}$ are set to 100, 1, 10, 10 respectively. Note that the length of the input and output sequences is set to 256 during training, but can be of any length during inference.


\begin{figure}
\centering
\includegraphics[width=0.49 \textwidth]{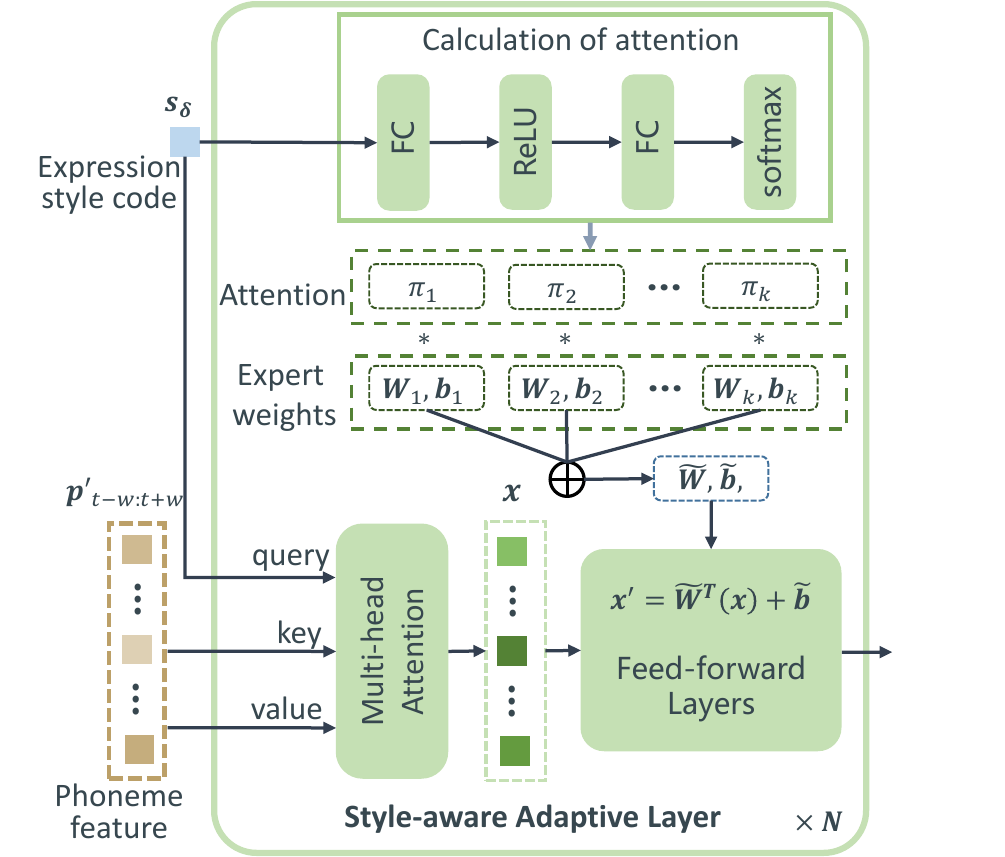}
\caption{Illustration of the style-aware adaptive transformer decoder layer.}
\label{fig:expression_decoder}
\end{figure}

\subsection{Stylized Facial Expression Generation}\label{style_expression_gen}
\subsubsection{Phoneme Encoder}
In the stylized facial expression generation module, we intend to merely extract articulation-related information from the audio. This will eliminate any interference that may affect the speaking style of the generated facial expressions, such as emotion and intensity. To achieve this, we utilize phoneme labels instead of acoustic features to represent the audio signals. The phoneme labels $p_{1:T}$ are subsequently transformed into phoneme embeddings and fed into the phoneme encoder $\mathbf{E}_p$, which produces sequential articulation representations $p'_{1:T}$, $p'_t \in \mathbb{R}^{256}$. The phoneme encoder comprises a vanilla transformer encoder, and the phoneme labels are extracted using a speech recognition tool.

\subsubsection{Style-Aware Facial Expression Decoder}
At the early stage, we employ the vanilla transformer decoder as the facial expression decoder $\mathbf{E}^{\delta}_d$, which takes the articulation representations $\boldsymbol{p}'_{t-w:t+w}$ and the facial expression style code $\boldsymbol{s}_{\delta}$ as input. Specifically, we repeat the expression style code $2w+1$ times and then add them with positional encodings to obtain the style tokens. The style tokens serve as the query of the transformer decoder, and the latent articulation representations serve as the key and value. The middle output token is fed into a feed-forward network to generate the output expression parameter $\boldsymbol{\hat{\delta}}_t$. The facial expression generation process can be formulated as:
\begin{equation}
    \boldsymbol{\hat{\delta}}_{1:T} = \mathbf{E}^{\delta}_d(\boldsymbol{p}'_{1:T},\boldsymbol{s}_{\delta})
\end{equation}

When utilizing the aforementioned decoder, we observe defective lip movements and facial expressions when generating stylized talking faces with large facial movements. Inspired by \cite{yang2019condconv} and \cite{karras2020analyzing}, we assume that the static kernel weights cannot model the  diverse speaking styles. With this assumption, we design a style-aware adaptive transformer, which dynamically adjusts the network weights according to the style code, as shown in Figure~\ref{fig:expression_decoder}. Specifically, since \cite{wang2020rethinking} reveals that the feed-forward layers play the most important role in transformer decoder, we replace the feed-forward layers with novel style-aware adaptive feed-forward layers. The style-aware adaptive layer utilizes $K=8$ parallel sets of weights ${\tilde{\boldsymbol{W}}_k,\tilde{\boldsymbol{b}}_k}$. Such parallel weights are expected to be the experts for modeling the distinct facial motion patterns of the different speaking styles. Then we introduce the additional layers followed by Softmax to adaptively compute the attention weights over each set of weights depending on the style code. Then the feed-forward layer weights are aggregated dynamically via the attention weights:
\begin{equation}
    \begin{aligned}
    \tilde{\boldsymbol{W}}(\boldsymbol{s}_{\delta}) &=\sum_{k=1}^{K} \pi_{k}(\boldsymbol{s}_{\delta}) \tilde{\boldsymbol{W}}_{k}, \tilde{\boldsymbol{b}}(\boldsymbol{s}_{\delta})=\sum_{k=1}^{K} \pi_{k}(\boldsymbol{s}_{\delta}) \tilde{\boldsymbol{b}}_{k}, \\\text { s.t. } & 0 \leq \pi_{k}(\boldsymbol{s}_{\delta}) \leq 1, \sum_{k=1}^{K} \pi_{k}(\boldsymbol{s}_{\delta})=1,
    \end{aligned}
\end{equation}
where $\pi_{k}$ is the attention weight for $k^{th}$ feed-forward layer weights ${\tilde{\boldsymbol{W}}_k,\tilde{\boldsymbol{b}}_k}$. The output of style-controllable dynamic feed-forward layers is then obtained by:
\begin{equation}
    \boldsymbol{y}=g\left(\tilde{\boldsymbol{W}}^{T}(\boldsymbol{s}_{\delta}) \boldsymbol{x}+\tilde{\boldsymbol{b}}(\boldsymbol{s}_{\delta})\right),
\end{equation}
where $g$ is an activation function. Our experiments show that the style-controllable dynamic decoder helps to create accurate stylized lip movements and natural stylized facial expressions in diverse speaking styles.

\subsubsection{Disentanglement of Upper and Lower faces}
In our experiments, we observed that the upper face and the lower face exhibit different motion patterns and have distinctive correlations with the audio input. Specifically, the upper face (eye, eyebrow) moves at a low frequency, while the lower face (mouth) moves at a high frequency. Thus, it is reasonable to model the motion patterns of the two parts with separate networks.

To begin, we divided the expression parameters into two groups: the lower face group and the upper face group. We then utilized two parallel style-controllable dynamic decoders, namely the upper face decoder and the lower face decoder, to generate the corresponding group of express parameters. For the lower face group, we selected 13 out of the 64 expression parameters that are highly related to mouth movements. For the upper face group, we used the remaining parameters. Finally, we concatenated the two groups of generated expression parameters to obtain the final generated expression parameters.

\subsubsection{Facial Expression Objective Function Design}
Because the stylized facial expression generation module generates each frame individually, we adopt a batched sequential training strategy~\cite{wang2022one} to improve the temporal consistency. Specifically, we generate successive $L=64$ frames $\boldsymbol{\delta}_{1:L}$ at one time as a clip. Specifically, we jointly train the phoneme encoder $\mathbf{E}_p$, the facial expression style encoder $\mathbf{E}^{\delta}_{s}$, and the style-aware facial expression decoder $\mathbf{E}^{\delta}_d$ using the following loss function design:


\begin{figure}[t]
\centering
\includegraphics[width=0.49 \textwidth]{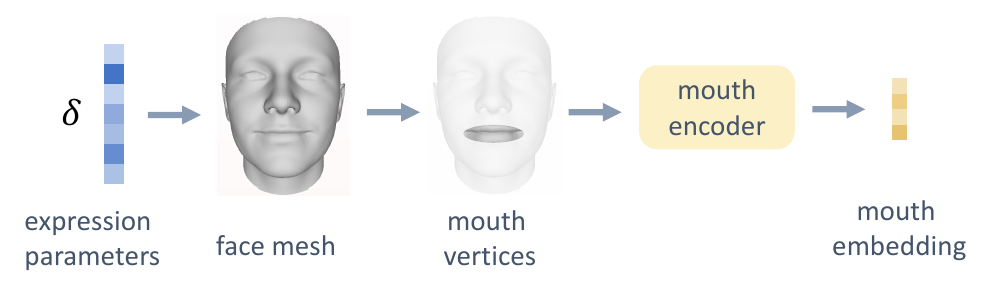}
\caption{Mouth embedding extraction in lip-sync discriminator.}
\label{fig:wav2lip}
\end{figure}

\newpara{Facial Expression Reconstruction Constraint:} During training, we reconstruct the facial expressions of each clip in the self-driven setting. We adopt a combination of the L1 loss and SSIM loss:
\begin{equation}
    \mathcal{L}^{\delta}_{rec} = \mu \mathcal{L}_{
    {L1}}(\boldsymbol{\delta}_{1:T},\boldsymbol{\hat{\delta}}_{1:T}) + (1-\mu) \mathcal{L}_{{ssim}}(\boldsymbol{\delta}_{1:T},\boldsymbol{\hat{\delta}}_{1:T}),
\end{equation}
where $\boldsymbol{\delta}_{1:T}$ and $\boldsymbol{\hat{\delta}}_{1:T}$ are the ground truth and reconstructed facial expressions respectively. $\mu$ is a ratio coefficient and is set to 0.1. 

\newpara{Lip-sync Discriminator.}
Due to the variability in mouth shape across different speaking styles, achieving accurate lip synchronization is an extremely challenging task. 
Inspired by SyncNet~\cite{prajwal2020lip}, we design a lip-sync discriminator $\mathbf{D}_{sync}$, which is trained to discriminate the synchronization between audio and mouth by randomly sampling an audio window that is either synchronous or asynchronous with a video window. 

\begin{table*}[!htbp] 
\centering
\caption{Quantitative results on the visual effects of the generated videos on MEAD and HDTF dataset.}
\setlength{\tabcolsep}{1.3mm}{
\begin{tabular}{ccccccccccc}
\toprule  
&\multicolumn{5}{c}{MEAD}&\multicolumn{5}{c}{HDTF} \\
\cmidrule(r){2-6}  \cmidrule(r){7-11}
Method & SSIM$\uparrow$ & CPBD$\uparrow$ & F-LMD $\downarrow$ & M-LMD $\downarrow$ &$\text{Sync}_{conf}$$\uparrow$ & SSIM$\uparrow$  & CPBD$\uparrow$ & F-LMD $\downarrow$  &   M-LMD $\downarrow$  & $\text{Sync}_{conf}$$\uparrow$ \\
\midrule

MakeitTalk       & 0.725 & 0.106 & 3.969 & 5.324 & 2.104 & 0.593 &  0.248 & 5.084 &  4.447 & 2.563  \\
Wav2Lip   & 0.795 & \textbf{0.178} & 2.718 & 4.052 & \textbf{5.257} & 0.618  & 0.299 & 4.544 & 3.630 & 3.072  \\
PC-AVS    & 0.504 & 0.071 & 5.828 & 4.970  & 2.183 & 0.422 & 0.132 & 10.506 & 3.931 & 2.701  \\
AVCT  & 0.832 &  0.139 & 2.923 & 5.520  & 2.525 & 0.755  &  0.233 & 2.733 & 3.610 & 3.147 \\
GC-AVT & 0.340  & 0.142 & 8.039 & 7.103 & 2.417 & 0.337 &  0.296 & 10.537 & 6.206  & 2.772  \\
EAMM & 0.397 & 0.084 & 6.698 & 6.478 & 1.405 & 0.387 &  0.144 & 7.031 & 6.857  & 1.799  \\
Ground Truth  & 1   &     0.222 & 0 & 0  & 4.131 & 1     &  0.307 & 0    & 0  & 3.961  \\

\cmidrule(r){1-11}
\textbf{Ours}  & \textbf{0.837} & 0.164 & \textbf{2.122}  & \textbf{3.249} & 3.474     & \textbf{0.812}   & \textbf{0.302} & \textbf{1.941}   & \textbf{2.412} & \textbf{3.165}  \\
\bottomrule 
\end{tabular}
}
\label{table:quantitive_evaluation}
\end{table*}

We have made modifications to the original SyncNet to enhance the lip sync of synthetic facial expressions. Since the 3DMM expression PCA bases controlling mouth movements also affect other facial movements, we first convert expression parameters into a face mesh using the PCA expression bases, and then extract the mouth vertices as a pure mouth shape representation, as illustrated in Figure~\ref{fig:wav2lip}. We specifically select 404 vertices located in the mouth area of face meshes in the 3D morphable model. We input the mesh vertex coordinates into the mouth encoder and phonemes into the audio encoder, instead of feeding images and acoustic features into the original SyncNet. 

Specifically, we use PointNet~\cite{qi2017pointnet} as the mouth encoder to extract the mouth embedding $\mathbf{e}_m$, and another phoneme encoder to compute the audio embedding $\mathbf{e}_a$ from the phoneme window. We adopt cosine similarity to indicate the probability that $e_m$ and $e_a$ are synchronous~\cite{prajwal2020lip}:
\begin{equation}
    P_{sync} =  \frac{\mathbf{e}_m \cdot \mathbf{e}_a}{\max({\| \mathbf{e}_m \|}_2 \cdot {\| \mathbf{e}_a \|}_2, \epsilon)},
\end{equation}
where $\epsilon$ is a small constant. The expression generation module maximizes the synchronous probability via a sync loss $\mathcal{L}_{sync }$ on each frame of the generated clip:
\begin{equation}
    \mathcal{L}_{sync} = \frac{1}{L} \sum_{i=1}^{L}-\text{log}(P_{sync}^i).
\end{equation}

\newpara{Facial Expression Style Discriminator.}
The facial expression style discriminator $\mathbf{D}^{\delta}_{style}$ is designed to classify the speaking style of the input sequential 3DMM expression parameters $\boldsymbol{\delta}_{1:L}$. Specifically, the style discriminator generates a probability distribution $P^s \in \mathbb{R}^{C}$, indicating the likelihood that the sequence of parameters belongs to each speaking style. Here, $C$ is the number of speaking styles. The style discriminator follows the PatchGAN structure, and is initially pre-trained on a dataset containing $C$ speaking styles using cross-entropy loss. Once pre-trained, the style discriminator is frozen and used to guide the generative modules towards producing vivid speaking styles via a style loss function $\mathcal{L}^{\delta}_{style}$:
\begin{equation}
    \mathcal{L}^h_{style} = -\text{log}(P_i^s),
\end{equation}
where $i$ is the category of the ground-truth speaking style of the facial expressions.

\newpara{Facial Expression Temporal Discriminator.}
To improve the temporal stability of the generated facial expressions, we utilize a facial expression temporal discriminator $\mathbf{D}^{\delta}_{tem}$, similar to the head pose temporal discriminator $\mathbf{D}^h_{tem}$, to calculate the adversarial loss $\mathcal{L}^{\delta}_{tem}$.

\newpara{Full Objective}
Our full objective for training the stylized facial expression generation module is given by a combination of the aforementioned loss terms:
\begin{equation}
\begin{split}
        \mathcal{L}^{\delta}_{total}=\lambda^{\delta}_{rec} \mathcal{L}^{\delta}_{rec}+\lambda^{\delta}_{trip} \mathcal{L}^{\delta}_{trip} + \lambda_{\text {sync }} \mathcal{L}_{\text {sync }} +\\
    + \lambda^{\delta}_{tem} \mathcal{L}^{\delta}_{tem}+\lambda^{\delta}_{style} \mathcal{L}^{\delta}_{style},
\end{split} 
\end{equation}
where $\mathcal{L}^{\delta}_{trip}$ is the triplet loss for the facial expression style code, as introduced in Section~\ref{style_encoder}, and we use $\lambda^{\delta}_{{rec }}=88$, $\lambda^{\delta}_{trip } = 1$, $\lambda_{sync } = 1$, $\lambda^{\delta}_{tem} = 1$ and $\lambda^{\delta}_{style } = 1$.

\subsection{Image Render}\label{renderer}
After obtaining the generated stylized head movements $\boldsymbol{\hat{h}}_{1:T}$ and stylized facial expressions $\boldsymbol{\hat{\delta}}_{1:T}$, we integrate them and feed them into an image renderer $\mathbf{E}_r$ along with the reference one-shot image to produce the final output videos. Our image renderer follows the network architecture of PIRenderer~\cite{ren2021pirenderer}, which is capable of generating photo-realistic results with accurate motions by utilizing a source portrait image and target 3DMM parameters. PIRenderer comprises a mapping network, a warping network, and an editing network. The mapping network produces latent vectors from the 3DMM parameters. Instructed by the latent vectors, the warping network estimates the dense motion field between the source and desired images, producing a coarse image with the estimated deformations. Finally, the editing network generates the final images from the coarse images.

\begin{figure*}[ht]
\centering
\includegraphics[width=0.98\textwidth]{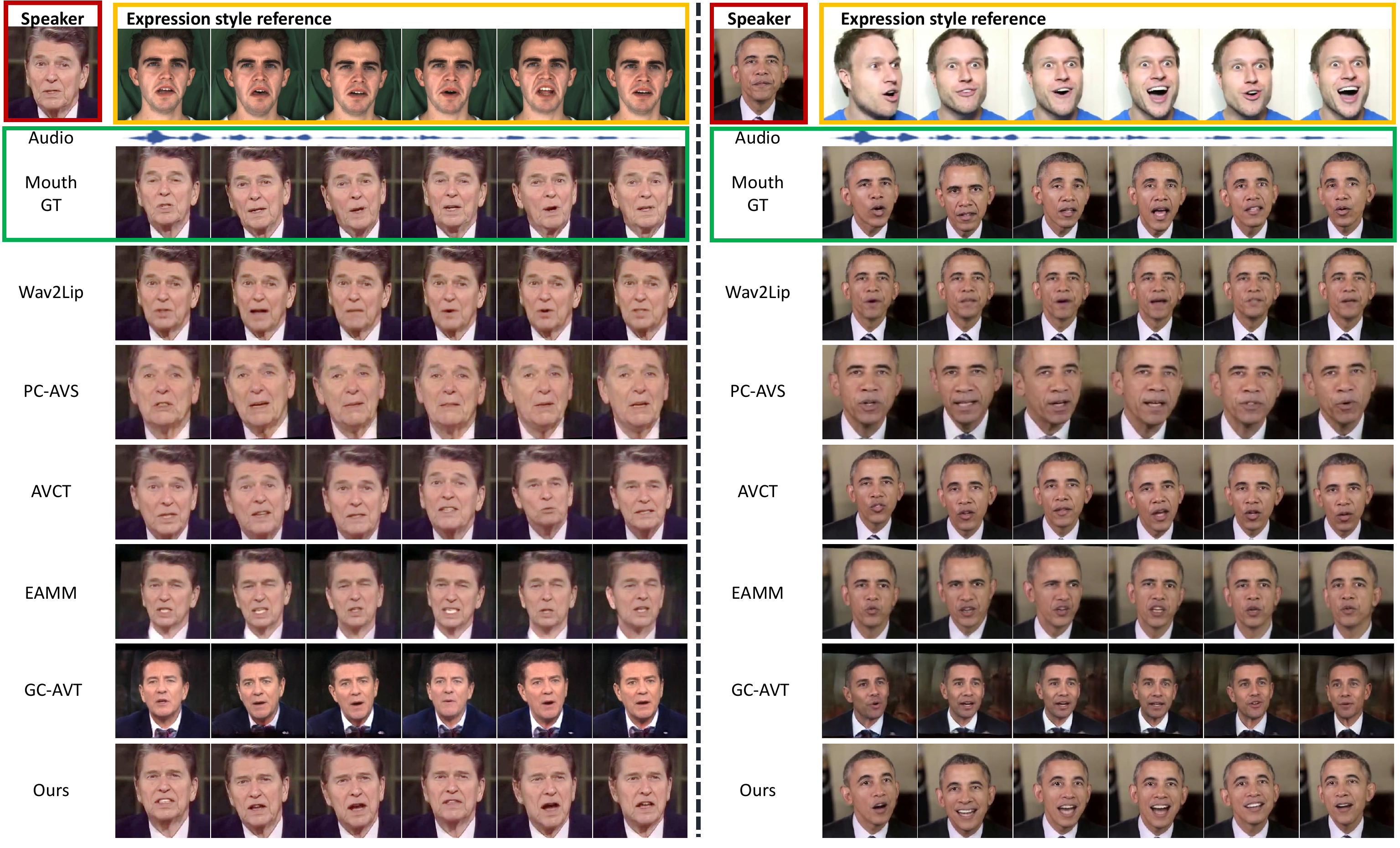}
\caption{Qualitative comparisons with the person agnostic methods. The identity reference, expression style reference videos, and audio-synced videos are displayed in the first two rows. This figure mainly showcases comparisons in visual quality, facial expression, and lip-sync accuracy. It is worth noting that for EAMM, GC-AVT, and our method, we use the same video clip as the expression style reference. For PC-AVS, AVCT, EAMM, GC-AVT, and our method, head poses are derived from the Mouth GT video. Please zoom in or see our demo video for more details.
}
\label{fig:qualitative}
\end{figure*}

\section{Implementations}

\subsection{Datasets}
In this paper, we use four widely-used talking face datasets: VoxCeleb~\cite{nagrani2017voxceleb}, HDTF~\cite{zhang2021flow}, MEAD~\cite{wang2020mead}, and HeadMotion~\cite{yi2022predicting}. All videos are aligned by cropping and resizing to $256\times256$, as done in \cite{siarohin2019first}. The videos are sampled at 25 FPS, and the audio is pre-processed to 16KHZ.

\newpara{VoxCeleb Dataset.} VoxCeleb is an audio-visual dataset that consists of short clips of human speech recorded in the wild. It comprises utterances from more than 1,000 speakers of different ethnicities, accents, professions, and ages. 

\newpara{HDTF Dataset.} HDTF consists of 362 high-quality videos of over 300 subjects. The resolution of original videos is 720P or 1080P. The test set comprises 20 videos, totaling around 10K frames.

\newpara{MEAD Dataset.} Mead is a high-quality emotional talking-face dataset recorded in the lab. It includes videos in which different speakers speak with eight different emotions at three different intensity levels. Here, we have selected 42 actors for training and 6 actors for testing. 

\newpara{HeadMotion Dataset.} HeadMotion is a recently built head motion dataset collected from the internet. This dataset consists of 751 single-person talking videos, each recorded without camera movements. Various types of head movements are contained within this dataset.



\subsection{Implementation Details}
\newpara{Style encoder.} The universal style encoder $\mathbf{E}_s$ takes as input the sequential expression parameters $\boldsymbol{\delta}_{1:N}$ or head poses $\boldsymbol{h}_{1:M}$. The length $N$ of the expression style reference video is from 64 to 256 (about $2 \sim 10$ seconds) and the length $M$ of the head pose style reference video is from 128 to 512 (about $5 \sim 20$ seconds). We increase their dimension to 256 by feeding them into a linear layer. Next, we feed the sequence features into a transformer encoder, which contains 3 8-head transformer encoder layers with a hidden size of 256. The output tokens are then aggregated by a self-attention pooling layer, as introduced in Section~\ref{style_encoder}, to obtain the final style code $\boldsymbol{s}$.

\newpara{Acoustic encoder and phoneme encoder.} Both $\mathbf{E}_a$ and $\mathbf{E}_p$ employ the same transformer encoder architecture as that used in the universal style encoder $\mathbf{E}_s$. In $\mathbf{E}_a$, the dimension of each frame's acoustic feature is first increased to 256 by a feed-forward layer. As phonemes are denoted as discrete labels, each phoneme label is mapped into a word embedding of 256 in $\mathbf{E}_a$.

\newpara{Head pose decoder.} We generate one frame of head pose in each step in the style-aware head pose decoder $\mathbb{E}^h_d$. At each time step $i$, the concatenated features of $\boldsymbol{a'_i}$, $\boldsymbol{e}_{i-1}$, and $\boldsymbol{s}_h$ are fed to a feed-forward layer to decrease the feature dimension to 256, and they are then appended to the end of the memory tokens. These tokens are fed to a 2-layer transformer, where the head number is 8 and the hidden size is 256. The last output token is used as $\boldsymbol{e}_i$, which is fed to a fully connected layer to produce the current head pose $\boldsymbol{h}_i$. The length of the memory is set to 128, and we truncate the last 128 tokens from the output tokens to form the new memory.

\newpara{Expression decoder.} Our style-aware expression decoder $\mathbb{E}^{\delta}_d$ is implemented based on the transformer decoder, where the style tokens serve as queries and the audio features serve as keys and values. The transformer decoder has three 8-head decoder layers and the hidden dimension is 256. In this decoder, we replaced the feed-forward layers with style-aware adaptive feed-forward layers. For these layers, we initialize eight sets of weights for a feed-forward layer with a hidden size of 2048. The style code is then fed into a linear attention network to obtain eight attention weights, which are used to aggregate the aforementioned weights. The attention network comprises two fully connected layers and a Softmax layer, with a hidden dimension of 64 for the two fully connected layers. Finally, the eight sets of weights are aggregated using the attention weights to obtain the final weights for the feed-forward layer.

\newpara{Image renderer.}  We used the well-known PIRenderer~\cite{ren2021pirenderer} as the network architecture for our image renderer. For detailed information on the architecture, please refer to \cite{ren2021pirenderer}. In this paper, we trained an off-the-shelf image renderer on three talking face datasets, namely HDTF~\cite{zhang2021flow}, VoxCeleb~\cite{nagrani2017voxceleb}, and MEAD~\cite{wang2020mead}.

\newpara{Temporal discriminator and style discriminator.} Our head pose temporal discriminator $\mathbf{D}^h_{tem}$, facial expression temporal discriminator $\mathbf{D}^{\delta}_{tem}$, head pose style discriminator $\mathbf{D}^h_{style}$, and facial expression style discriminator $\mathbf{D}^{\delta}_{style}$ have similar network architectures. We replaced the 2D convolutions in the vanilla $70 \times 70$ PatchGAN~\cite{isola2017image} with 1D convolutions and performed convolution operations along the temporal dimension. The discriminator judges whether each 70-length input sequence is real or fake. 



\begin{table}
\centering
\caption{Quantitative results of generated head movements on HDTF and HeadMotion dataset.}
\begin{tabular}{ccccc}
\toprule  
&\multicolumn{2}{c}{HDTF}&\multicolumn{2}{c}{HeadMotion} \\
\cmidrule(r){2-3}  \cmidrule(r){4-5}
Method & SSIM$\uparrow$ & PSNR$\uparrow$ &  SSIM$\uparrow$ & PSNR$\uparrow$\\
\midrule
MakeitTalk & 0.747 & 25.49 & 0.646 & 24.66 \\
Audio2Head & 0.761 & 26.52 & 0.707 & 26.11 \\

\cmidrule(r){1-5}
Ours & \textbf{0.847} & \textbf{27.58} & \textbf{0.786} & \textbf{26.75} \\
\bottomrule 
\end{tabular}

\label{table:quantitive_head_pose}
\end{table}

\subsection{Training Details}
We use PyTorch~\cite{paszke2017automatic} to implement our method.
Our framework is implemented by Pytorch~\cite{paszke2017automatic}. We employ Adam optimizer \cite{kingma2014adam} for all training. 

The branch responsible for generating stylized head poses jointly trains models $\mathbf{E}^h_s$, $\mathbf{E}_a$, $\mathbf{E}^h_d$, and $\mathbf{D}^h_{tem}$ with an initial learning rate of 1e-4, which decays to 2e-6 within 100 epochs. Specifically, during the first 50 epochs, $\mathbf{D}^h_{style}$ is not involved in the training, but is then jointly trained with other modules in the following 50 epochs. These models are trained on a combination of the HDTF and HeadMotion datasets. We observe that individuals maintain a consistent head movement style over time. Therefore, we sample triplet-paired samples according to the following strategy for the triplet constraint (Section. \ref{style_encoder}) of the head pose style encoder during training. Specifically, head pose style video clips sampled from segments near the anchor are considered to have the same style as the anchor, while clips sampled from non-adjacent segments have a different style. 

To train the modules in the stylized facial expression generation branch, we first construct our dataset based on MEAD and HDTF. For MEAD, we assume that video clips where the speaker expresses the same emotion at the same intensity level share the same expression style. For HDTF, we assume that video clips from one speaker share the same speaking style. We obtain 1,104 speaking styles, and each style corresponds to a set of videos in the training set. For the triplet constraint of the expression style encoder, positive samples are sampled from the same set as the given one, while negative samples are sampled from any other set. $\mathbf{D}_\text{sync}$ and $\mathbf{D}^{\delta}_\text{style}$ are trained on this dataset with a learning rate of 0.0001. Afterward, $\mathbf{D}_\text{sync}$ and $\mathbf{D}^{\delta}_\text{style}$ are frozen, and $\mathbf{E}_a$, $\mathbf{E}_s$, $\mathbf{E}_d$, and $\mathbf{D}_\text{tem}$ are jointly trained for 50 epochs with a learning rate of 0.0001.

\subsection{Metrics}
In this paper, we use several widely adopted metrics to evaluate the effectiveness of the proposed methods. For evaluating lip synchronization, we use the confidence score of SyncNet~\cite{chung2016out} ($\textbf{Sync}_{\textbf{conf}}$) and the Landmark Distance around the mouth (\textbf{M-LMD}) \cite{chen2019hierarchical}. To assess the accuracy of generated facial expressions, we use the Landmark Distance on the whole face (\textbf{F-LMD}). To evaluate the quality of generated talking head videos, we adopt \textbf{SSIM} and the Cumulative Probability of Blur Detection (\textbf{CPBD}) \cite{narvekar2009no}. Furthermore, we separately evaluate the quality of the generated head pose using \textbf{SSIM} and peak signal-to-noise ratio (\textbf{PSNR}).


\section{Experiments}
In this section, we perform extensive experiments to validate our proposed method. We conduct quantitative evaluations with SOTA in Section~\ref{quanlitative_veal} and qualitative evaluations in Section~\ref{quantitative_eval}. In Section~\ref{ablation_eval}, we conduct ablation studies to validate the effectiveness of each component in our stylized head pose generation branch and stylized facial expression generation branch. We also conduct experiments in Section~\ref{style_space} to inspect the learned style code space. Furthermore, in Section~\ref{user_study}, we perform a user study. Finally, in Section~\ref{discussion}, we provide a discussion. To present and compare the results more clearly, most of the experiments in this section are divided into two groups to assess the overall visual effects and the head motion effects separately. Specifically, for evaluating the visual effects, we fix the head motion style, while for evaluating the head motion effects, we fix the facial expression style. 
\subsection{Quantitative Evaluation}\label{quantitative_eval}
To comprehensively evaluate our method, we conduct two sets of quantitative experiments. The first set focuses on evaluating the visual quality and facial expressions, including lip sync, of the generated videos. The second set focused on evaluating the quality of the generated head poses.

\begin{figure}
\centering
\includegraphics[width=0.48\textwidth]{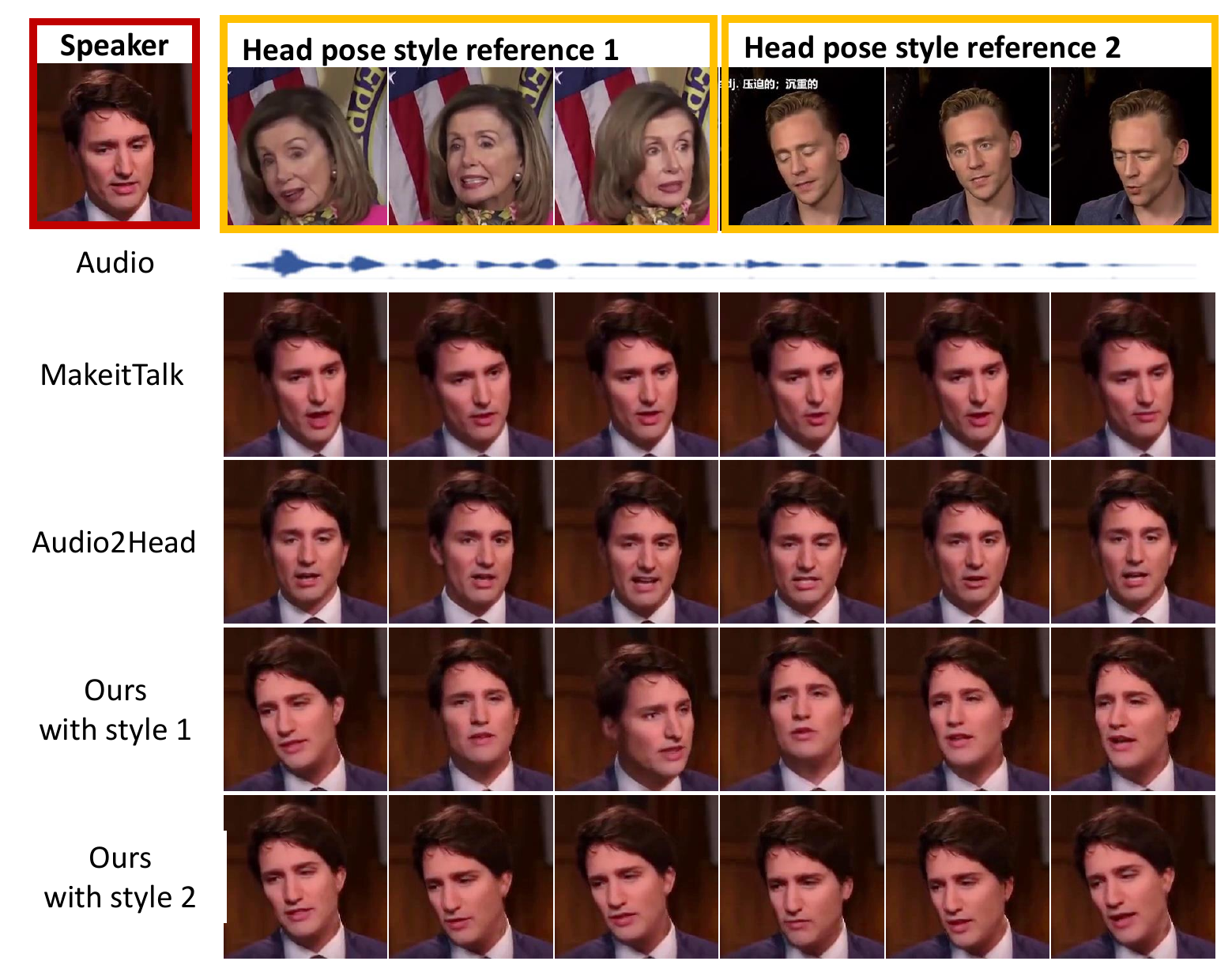}
\caption{Qualitative comparisons with the person agnostic methods that are capable of synthesizing head poses from audio. This figure primarily demonstrates comparisons of generated head movements. For our method, we use two video clips as the head pose style references. Style 1 presents a motion pattern of frequent left and right head shaking, while Style 2 mainly looks to the right when speaking.
}
\label{fig:qualitative_headpose}
\end{figure}

\begin{figure}
\centering
\includegraphics[width=0.49 \textwidth]{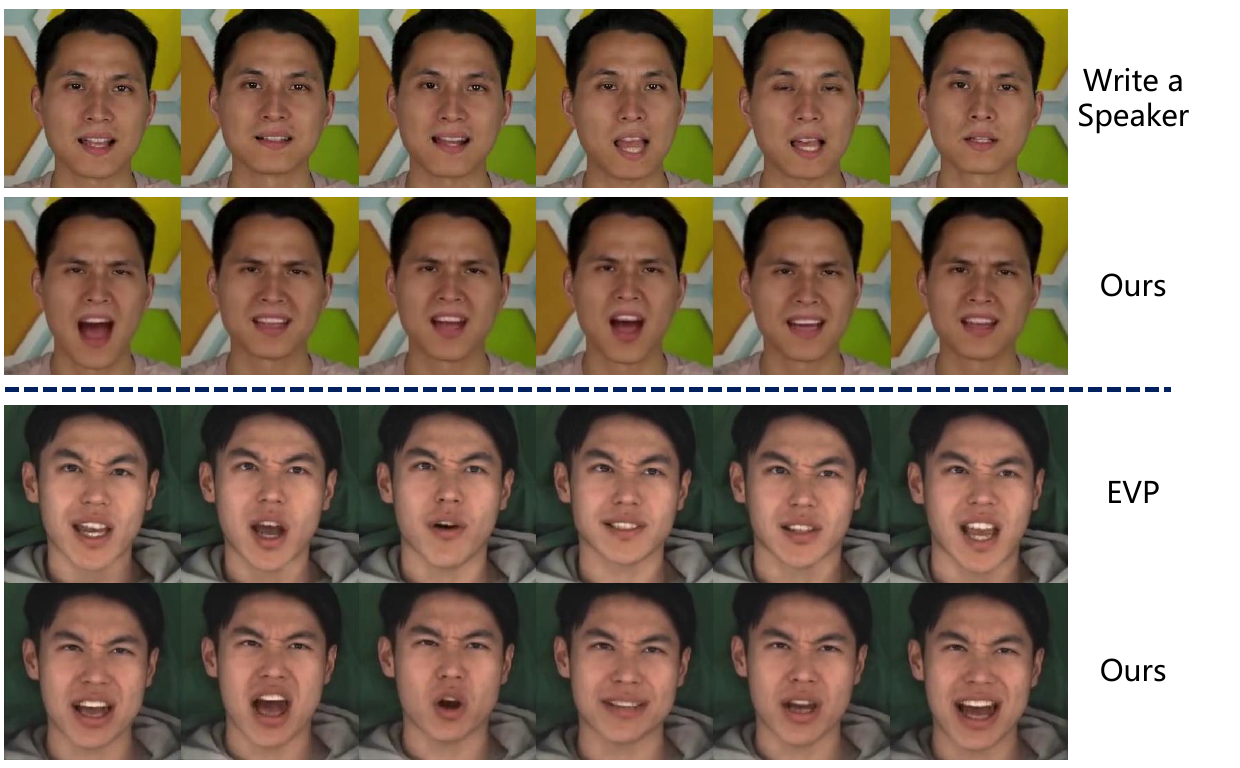}
\caption{Qualitative comparisons with person-specific methods.}
\label{fig:qualitative_specific}
\end{figure}

\subsubsection{Evaluation on Visual Effect}
We first conduct quantitative evaluations on the overall quality of the generated talking videos. We compare our method with state-of-the-art talking face methods, including MakeitTalk~\cite{zhou2020makelttalk}, Wav2Lip~\cite{prajwal2020lip}, PC-AVS~\cite{zhou2021pose}, AVCT~\cite{wang2022one}, GC-AVT~\cite{liang2022expressive}, and EAMM~\cite{ji2022eamm}. The experiments are performed in the self-driven setting on the test set of MEAD and HDTF, where the speaker and speaking style are not seen during training. We select the first image of each video as the reference image and use the corresponding audio clip as the audio input. For all methods, poses are derived from ground truth videos. However, Wav2Lip can only generate mouth area movements, so the head poses are fixed in its samples. EAMM, GC-AVT and our method require an additional expression reference video as input. For these methods, the ground truth videos are used as the expression reference videos. The compared methods' samples are generated using either their released codes or with the help of their authors. Table \ref{table:quantitive_evaluation} reports the results of the quantitative evaluation.

Our method outperforms most other methods in terms of various metrics on both MEAD and HDTF datasets. Since Wav2Lip merely generates mouth movements and does not change other parts of the reference images, it obtains the highest CPBD score on MEAD. However, the mouth area generated by Wav2Lip is blurry (See Figure \ref{fig:qualitative}). Since Wav2Lip is trained using SyncNet as a discriminator, it is reasonable for Wav2Lip to obtain the highest confidence score of SyncNet (${Sync}_{{conf}}$) on MEAD. The score is even higher than that of the ground truth. Our method achieves the closest $\text{Sync}_{\text{conf}}$ scores to ground truth on MEAD and the highest on HDTF, indicating our method's ability to produce precise lip-sync. In terms of M-LMD metric, our method achieves the best scores, further demonstrating the accuracy of our lip-sync generation.  Moreover, our method performs the best under the F-LMD metric, demonstrating our method's ability to generate facial expressions that match the reference speaking style. Therefore, our method outperforms other methods in generating high-quality lip-sync with accurate facial expressions across various metrics.


\subsubsection{Evaluation on Head Movements}
We then conduct another quantitative evaluation on the generated head poses by comparing our method with MakeitTalk, Audio2Head~\cite{wang2021audio2head}, and PHM~\cite{yi2022predicting}. These methods are capable of generating head movements from audio. The experiments are also performed in the self-driven setting on the test sets of HDTF and HeadMotion. In each case, we selected the first image of each video as the reference image and used the corresponding audio clip as the input. For PHM and our method, we used the ground truth video as the source of the head pose style. For MakeitTalk, Audio2Head, and PHM, we extract the head pose sequence from their generated videos. As can be seen in Table~\ref{table:quantitive_head_pose}, our method achieved the best scores on all metrics on HDTF and HeadMotion datasets.

\subsection{Qualitative Evaluation}\label{quanlitative_veal}
In addition to the person-agnostic methods, we also conduct qualitative evaluations using person-specific methods to demonstrate the superiority of our proposed method.

\subsubsection{Comparison with Person-Agnostic Methods}
We conduct two sets of qualitative evaluations to assess the overall visual effect and generated head poses independently when compared to person-agnostic methods. In the first set, we compare our method with speaker-agnostic (one-shot) methods, including Wav2Lip, PC-AVS, AVCT, EAMM, and GC-AVT, to visualize the video quality, lip-sync, and facial expression. Figure~\ref{fig:qualitative} displays the results of this comparison. Note that for EAMM, GC-AVT, and our method, we use the same video clip as the expression style reference. For PC-AVS, AVCT, EAMM, GC-AVT, and our method, we extract the head poses from the Mouth-GT videos where the audio input comes from. For the second set of qualitative evaluations, we conduct a comparison with Audio2Head and MakeitTalk to visualize the generated head poses. Figure~\ref{fig:qualitative_headpose} displays the results of this comparison, where we selected a neutral talking video as the expression style reference for our method. It is important to note that in both sets of evaluations, the identity reference, style reference, and audio were all unseen during training.


As shown in Figure~\ref{fig:qualitative}, our method can generate talking faces that accurately match a reference expression style, while achieving precise lip-sync and better preservation of the speaker's identity (please refer to our demo video). Among all the methods, only EAMM, GC-AVT, and our method can perform expression style control. However, EAMM and GC-AVT can only control expression styles in the upper face, such as the eyes and eyebrows, while failing to control the stylized shape of the mouth. Moreover, the expression styles of videos generated by these methods are significantly inconsistent with those of the style reference. In terms of lip-sync, only Wav2Lip, AVCT, PC-AVS, and GC-AVT are competitive with our method. However, they only model one neutral speaking style in the mouth area, making them unable to produce natural lip-sync in various styles. Furthermore, GC-AVT is unable to preserve the speaker's identity well, and both EAMM and GC-AVT are incapable of producing realistic backgrounds. In contrast, our method can imitate speaking styles in the entire face from arbitrary style clips while achieving accurate lip-sync, preserving speaker identity, and generating plausible backgrounds.



Based on Figure~\ref{fig:qualitative_headpose}, it is evident that our method is capable of extracting distinct head motion patterns from the reference video and generating diverse head motion sequences using the same audio input, under the guidance of the extracted head motion patterns. Furthermore, the different head pose sequences generated are synchronized with the rhythm of the same input audio. However, Audio2Head can only generate natural head movements and cannot control the style of the synthesized head motions. As for MakeItTalk, it can only generate slight movements that swing around the initial head pose in the reference image. Another noteworthy aspect is that we are able to apply head motion styles obtained from different camera planes to the reference image in its own plane of space.  For example, in Figure~\ref{fig:qualitative_headpose}, the speaker in the \textit{head pose reference style 2} is farther away from the camera than the subject in the reference image, and our method predicts the stylized head movements from the perspective of the reference image camera. This also indicates that our method indeed extracts head motion patterns from the reference video rather than simply transferring head movements.

\subsubsection{Comparison with Person-Specific Methods}
We further compare our method with person-specific emotional talking face methods, including Write-a-Speaker~\cite{li2021write} and EVP~\cite{ji2021audio}. For both methods, we crop video clips from their demo videos. Then we select one neutral image as the reference image and a video in MEAD with the same emotion as the style clip. The qualitative results are shown in Figure \ref{fig:qualitative_specific}. Compared with the other two works, our method also generates vivid emotional facial expressions and achieves comparable lip-sync. Note that our method is based on the one-shot setting, while the other two methods are trained on a long reference video of the target speaker.

\begin{table}[t]
\caption{ Quantitative results of the ablation study on the stylized head pose generation branch on the HeadMotion dataset.}
\centering
\setlength{\tabcolsep}{5mm}{
\begin{tabular}{ccc}
\toprule  
Method & SSIM$\uparrow$ &  PSRN$\uparrow$ \\
\midrule
\revision{w/ SDT-TF} & 0.566 & 23.56 \\
\revision{w/ SDT-RR} & \revision{0.643} & \revision{24.86} \\
w/o SET & 0.629 & 24.15 \\
w/o $\mathbf{D}^h_{style}$ & 0.736 & 25.78 \\
w/o $\mathcal{L}^h_{trip}$ & 0.683 & 25.10 \\
w/o $\mathbf{D}^h_{tem}$ & 0.710 & 25.52 \\
\revision{w/ Ph} & \revision{0.593} & \revision{24.22} \\
\cmidrule(r){1-3} 
\textbf{Full} & \textbf{0.786}  & \textbf{26.75} \\
\bottomrule 
\end{tabular}}

\label{table:Ablation_study_headpose}
\end{table}

\begin{table}[t]
\small
\caption{ Quantitative results of the ablation study on the stylized expression generation branch on the MEAD dataset.}
\centering
\setlength{\tabcolsep}{0.5mm}{
\begin{tabular}{cccccc}
\toprule  
Method & SSIM$\uparrow$ &  CPBD$\uparrow$ & F-LMD $\downarrow$  & M-LMD $\downarrow$ & $\text{Sync}_{conf}$$\uparrow$ \\
\midrule
\revision{w/o StyQ}    & \revision{0.836} &  \revision{0.161} & \revision{2.403} & \revision{3.651} & \revision{3.455} \\
w/o DyFFN    & 0.830 &  \textbf{0.165} & 2.414 & 4.178 & 3.059 \\
$K=4$      & 0.831  & 0.163 & 2.327 & 3.524 & 3.331 \\
$K=16$  & 0.835  & 0.161 & 2.133 & 3.396 & 3.473 \\
w/o $\mathbf{D}^{\delta}_{style}$    & 0.836  & 0.160 & 2.483   & 3.628 & 3.430  \\
w/o $\mathcal{L}^{\delta}_{{trip}}$   & {0.837}  & 0.160 & 2.401 &  3.771 & \textbf{3.532} \\
w/o $\mathbf{D}_{{sync}}$  & 0.834  & 0.164 & 2.281 & 4.351 & 2.305\\
\cmidrule(r){1-6}
\textbf{Full ($K=8$)}            & \textbf{0.837}  & 0.164 & \textbf{2.122}  & \textbf{3.249} & 3.474 \\
\bottomrule 
\end{tabular}}

\label{table:Ablation_study}
\end{table}

\subsection{Ablation Study}\label{ablation_eval}
\subsubsection{Ablation Study on Stylized Head Pose Generation}
\begin{figure}
\centering
\includegraphics[width=0.48\textwidth]{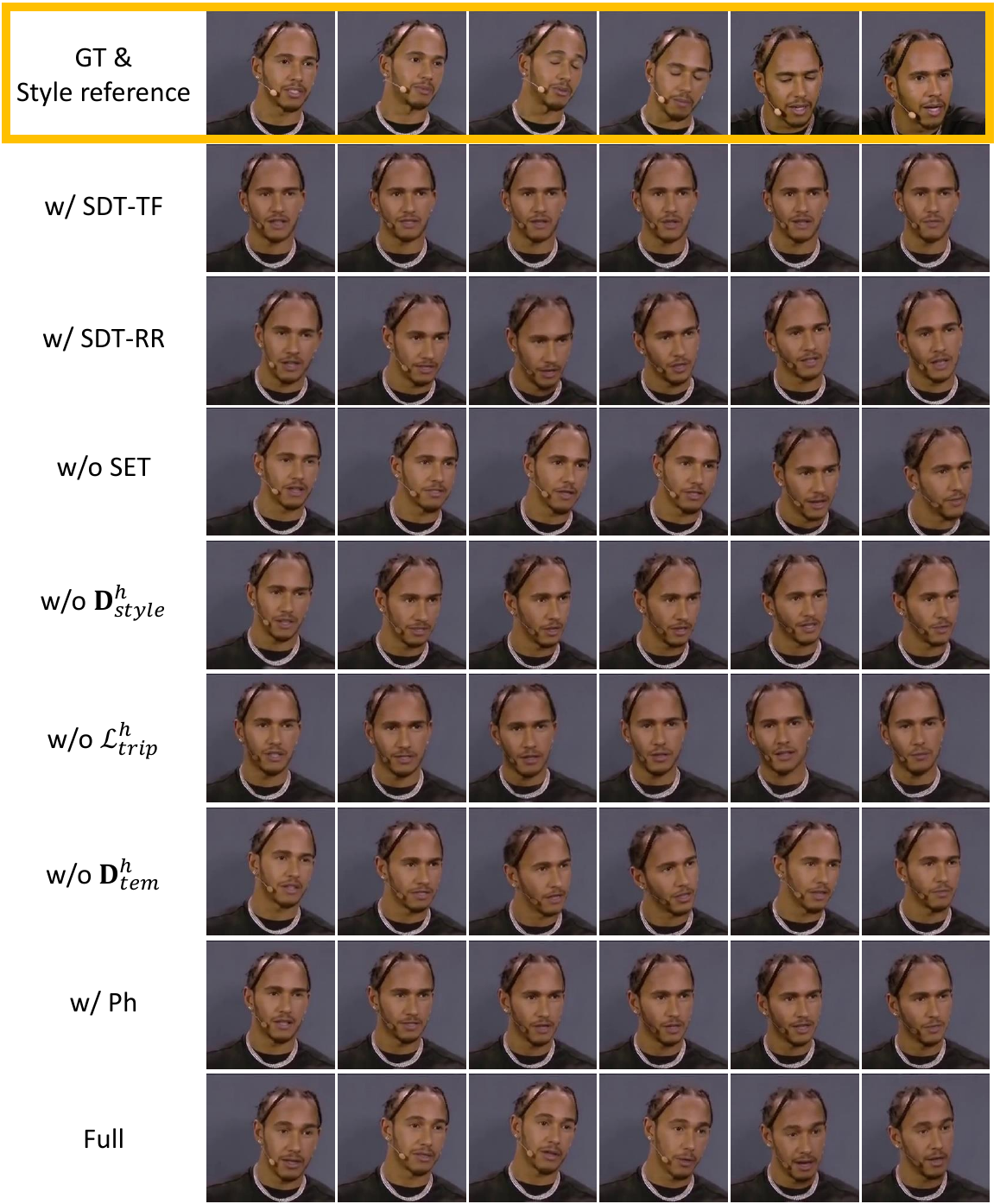}
\caption{Qualitative results of the ablation study on stylized head pose generation branch.
}
\label{fig:ablation_headpose}
\end{figure}

We first conduct ablation studies on the components in the stylized head pose generation modules on the HeadMotion dataset. Specifically, we analyze the impact of removing individual modules from the overall system. We design eight variants: (1) replace the Transformer-XL with the standard transformer decoder (\textbf{\revision{w/ SDT-TF}}) and \revision{train the model with teacher-forcing strategy}, \revision{(2) replace the Transformer-XL with the standard transformer decoder (\textbf{\revision{w/ SDT-RR}}) and train the model using recurrence strategy}, (3) remove the spatial embedding transition by eliminating the input of $\boldsymbol{e}_{i-1}$ at each time step $i$ (\textbf{w/o SET}),  (4) remove the style discriminator (\textbf{w/o $\mathbf{D}^h_{style}$}), (5) remove triplet loss (\textbf{w/o $\boldsymbol{\mathcal{L}}^h_{{trip}}$}, note that we also remove $\mathbf{D}^h_{style}$ in this variant), (6) remove the temporal discriminator (\textbf{w/o $\mathbf{D}^h_{tem}$}), \revision{(7) use phonemes as input instead of acoustic features (\textbf{w/ Ph})},  and (8) our full model (\textbf{Full}).  The results are shown in Table~\ref{table:Ablation_study_headpose} and Figure~\ref{fig:ablation_headpose}. Please see our demo video for more details on the dynamics of the generated head movements.

\begin{figure}
\centering
\includegraphics[width=0.49 \textwidth]{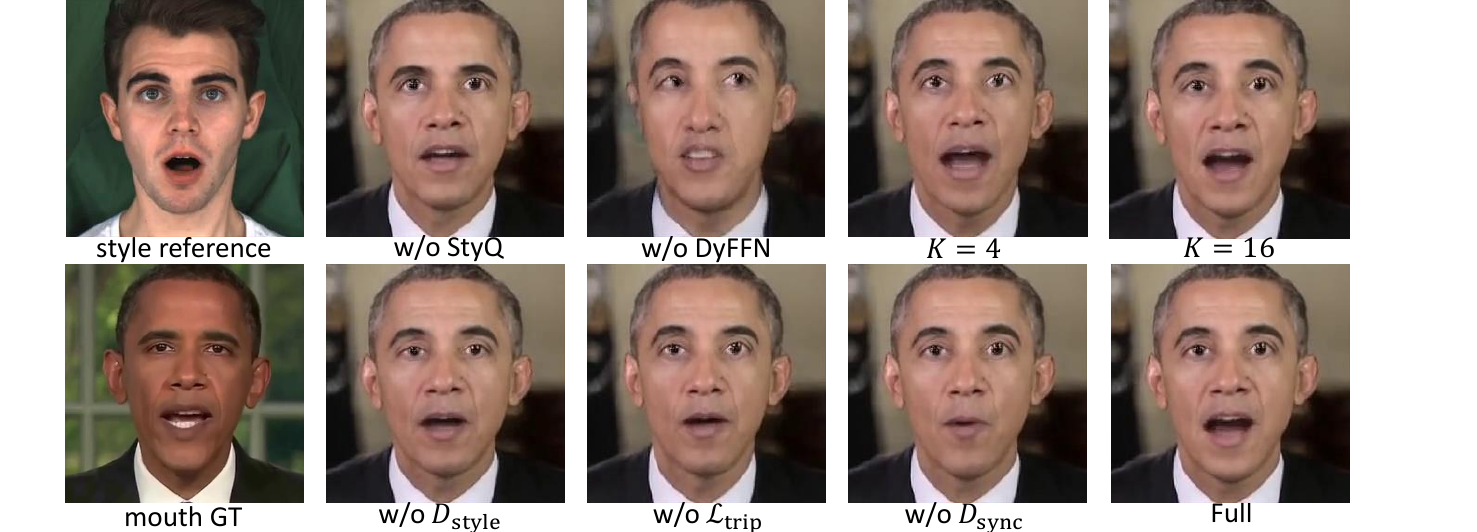}
\caption{Qualitative results of the ablation study on stylized expression generation branch.}
\label{fig:Qualitative_ablation}
\end{figure}

An interesting observation is that when we replace the Transformer-XL with the standard transformer (\textbf{w/ SDT-TF}), the network tends to only produce minimal head movements around the initial position. indicating that the recurrence is essential for enhancing the dynamism of head movement generation. Furthermore, \revision{despite both \textbf{w/ SDT-TF} and \textbf{w/o SET} applying the recurrence strategy, neither explicitly inputs the spatial head pose embedding from the previous moment to the current step, resulting in highly unstable head movements.} This implies that subsequent movements depend on both the current speech and the current head movements. Even though both \textbf{w/o $\mathbf{D}^h_{style}$} and \textbf{w/o $\mathbf{D}^h_{style}$} can produce natural head movements, they achieve lower scores than our full model. This suggests that the triplet constraint and the style discriminator make our model sensitive to head motion patterns. Without using $\mathbf{D}^h_{tem}$, the generated head movements will show jitters. \revision{Moreover, when we replace the input acoustic features with phonemes, our model tends to produce head movements with smaller dynamics and loses rhythm corresponding with audio. This implies that incorporating acoustic features enhances the performance of the generated head motion.
Therefore, these results indicate that each component in the stylized head pose generation modules contributes significantly to the improvements of the final results.}

\begin{figure*}[ht]
\centering
\includegraphics[width=0.98 \textwidth]{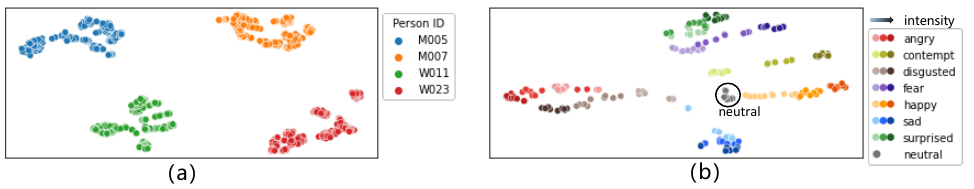}
\caption{(a) Visualization of the expression style codes of four speakers in MEAD. (b) Visualization of the emotional expression style codes of the speaker W011 in MEAD. Darker colors indicate higher emotion intensity.}
\label{fig:tsne_all}
\end{figure*}

\begin{figure}[ht]
\centering
\includegraphics[width=0.48 \textwidth]{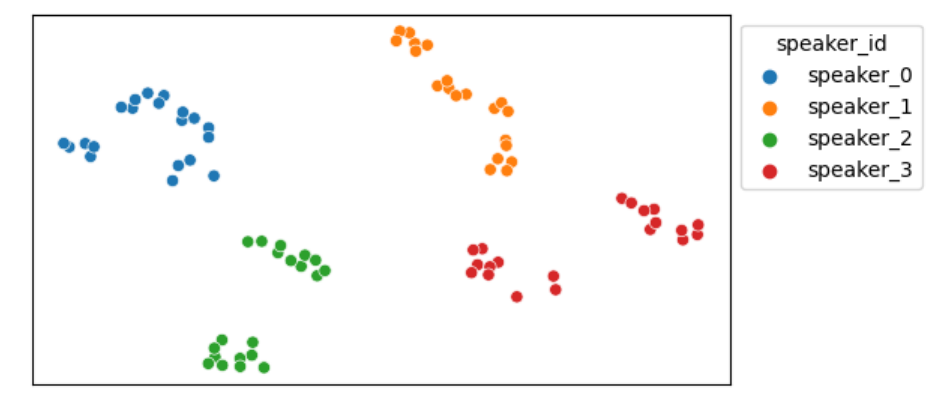}
\caption{Visualization of the head pose style codes of four speakers in HeadMotion dataset.}
\label{fig:tsne_head}
\end{figure}

\subsubsection{Ablation Study on Stylized Facial Expression Generation Modules}
We then conduct ablation studies on the components in the stylized facial expression generation modules on the MEAD dataset. Similarly, we analyze the impact of removing individual modules from the overall system or altering some settings. We design eight variants: \revision{(1) use the phoneme representation as the query in the decoder, rather than the style code (\textbf{w/o StyQ}),} (2) replace the adaptive feedforward layer with the vanilla feedforward layer (\textbf{w/o DyFFN}), (3) set $\boldsymbol{K}=4$ in dynamic feedforward layer ($\boldsymbol{K=4}$), (4) set  $\boldsymbol{K}= 16$ in dynamic feedforward layer ($\boldsymbol{K=16}$), 
(5) remove the style discriminator $\mathbf{D}^{\delta}_{style}$ (\textbf{w/o $\mathbf{D}^{\delta}_{{style}}$}),
(6) remove triplet loss (\textbf{w/o $\boldsymbol{\mathcal{L}}^{\delta}_{{trip}}$}),
(7) remove the lip-sync discriminator $\mathbf{D}_{{sync}}$ (\textbf{w/o $\mathbf{D}_{{sync}}$}), 
and (8) our full model (\textbf{Full}). The results are shown in Table \ref{table:Ablation_study} and Figure \ref{fig:Qualitative_ablation}.

Since all variants utilize the same image renderer, they obtain similar SSIM and CPBD scores. \revision{Compared to \textbf{Full}, both \textbf{w/o DyFFN} and \textbf{w/o StyQ} achieve lower scores in F-LMD, M-LMD, and $\text{Sync}_{conf}$.
Results in Figure \ref{fig:Qualitative_ablation} reveal that without using the style code as a query (\textbf{w/o StyQ}), stable facial expressions and good lip-sync can still be realized; however, the consistency between the generated and reference expressions decreases significantly. On the other hand, While \textbf{w/o DyFFN} generally produces animations that maintain consistent expressions with the reference style clip, it sometimes leads to unstable facial animations (see Figure~\ref{fig:Qualitative_ablation}). Therefore, using the style code as a query enhances the consistency between the synthesized and reference expressions, while using the adaptive feed-forward layer improves the stability of synthesized expressions and the accuracy of the mouth shape under various styles.}

We empirically observe that \textbf{$K=8$} is the optimal setting for our task. Without $\mathbf{D}^{\delta}_{{style}}$ and $\boldsymbol{\mathcal{L}}^{\delta}_{{trip}}$, the F-LMD and M-LMD scores drop dramatically. This implies that the style discriminator and the triplet constraint compel our framework to better perceive the stylized facial motion patterns. Furthermore, the results show poor lip synchronization when $\mathbf{D}_{{sync}}$ supervision is not included. \revision{Figure \ref{fig:Qualitative_ablation} and our demo video more clearly demonstrate the improvement each component brings to the final results.}



\begin{figure}
\centering
\includegraphics[width=0.49 \textwidth]{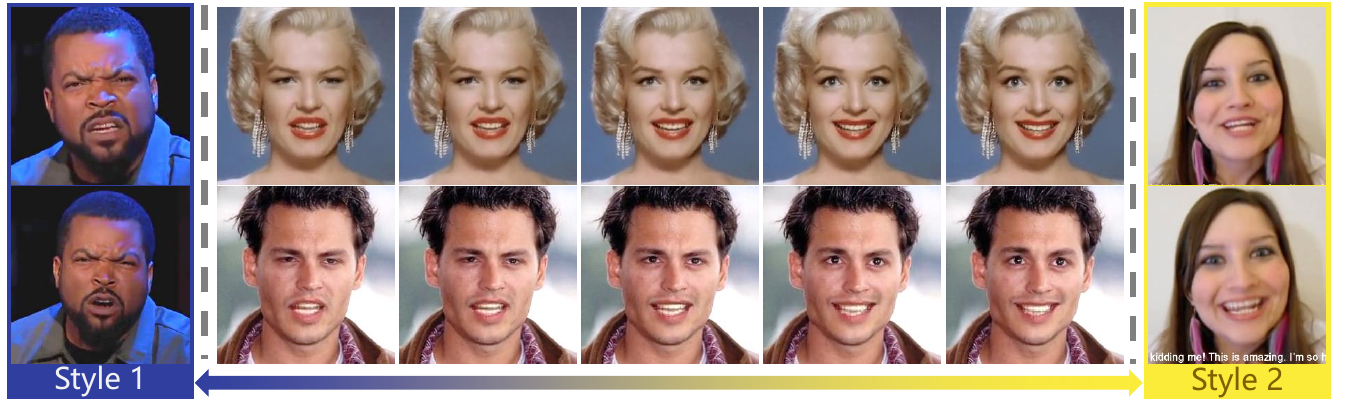}
\caption{Interpolation results between 2 expression styles. For each interpolated style, we show the results of 2 reference speakers.}
\label{fig:interpolation}
\end{figure}

\begin{figure}
\centering
\includegraphics[width=0.49 \textwidth]{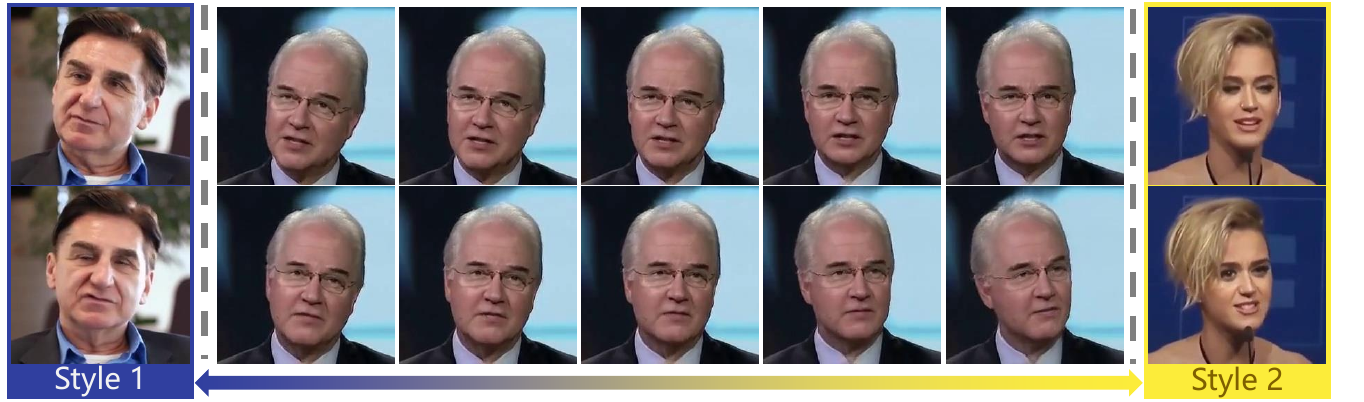}
\caption{Interpolation results between 2 head pose styles. For each style, we generate a video, and each video showcases two frames as examples.}
\label{fig:interpolation_pose}
\end{figure}

\subsection{Style Space Inspection}\label{style_space}
\subsubsection{Style Space Visualization}
For ease of visualization, We project the style codes to a 2D space using t-distributed stochastic neighbor embedding (t-SNE) \cite{van2008visualizing}.

To visualize the expression style codes, we selected four speakers from the MEAD dataset. Each speaker had 22 expression styles (7 emotions x 3 levels plus one neutral style). For each style, we randomly selected 10 video clips to extract style codes. In Figure \ref{fig:tsne_all}(a), each speaker is marked with a distinct color. As shown, the style codes of the same speaker cluster in the style space. This implies that the expression styles of one speaker are more similar to each other than to those of the same emotion from other speakers. Figure \ref{fig:tsne_all}(b) shows the style codes from one speaker in the MEAD dataset. Each style code is marked with a color corresponding to its emotion and intensity. Each group of style codes with the same emotion gathers into one cluster. In each cluster, the style codes of emotions with low intensity are close to those of the neutral emotion. Notably, some emotions show similar facial motion patterns, such as anger vs disgust and surprise vs fear. Thus, their style codes are close in the style space.

We selected 4 video clips (30-60 seconds) of 4 speakers from the HeadMotion dataset and randomly cropped 20 clips of 5-10 seconds from each video clip to extract the head pose style codes. In Figure~\ref{fig:tsne_head}, we can see that the style codes of the same speaker lie in nearby space, demonstrating that our model can accurately capture individual head pose styles.

The aforementioned observations prove that our model is able to learn a semantically meaningful style space.

\begin{table*}
\centering
\caption{Results of the user study on visual effects. The scores range from 1 to 5. Large scores indicate better perception. Here, the average scores across 24 videos are reported. All video are generated using the same input.}
\setlength{\tabcolsep}{4.3mm}{
\begin{tabular}{cccccccc}
\toprule  
Method & Wav2Lip & PC-AVS & AVCT & EAMM & GC-AVT & Ground Truth & Ours \\
\midrule
Lip Sync & \revision{3.45} & \revision{3.23} & \revision{3.24} & \revision{1.89} & \revision{3.41} & \revision{4.47} & \revision{\textbf{3.52}} \\
Video Realness & \revision{1.67} & \revision{2.03} & \revision{2.74} & \revision{1.39} & \revision{1.43} & \revision{4.24} & \revision{\textbf{3.06}} \\
Expression style Consistency & \revision{1.22} & \revision{1.68} & \revision{1.62} & \revision{1.80} & \revision{2.65} & \revision{1.85} &\revision{\textbf{3.46}} \\
\bottomrule 
\end{tabular}}
\label{table:user_study}
\end{table*}

\begin{table*}
\centering
\caption{Results of the user study on the generated head movements.}
\setlength{\tabcolsep}{5mm}{
\begin{tabular}{ccccc}
\toprule  
Method & MakeitTalk & Audio2Head & Ground Truth & Ours \\
\midrule
Head motion naturalness & \revision{1.66} & \revision{3.41} & \revision{4.91} & \revision{\textbf{3.52}} \\
Head motion sync & \revision{2.12} & \revision{\textbf{3.94}} & \revision{4.86} & \revision{3.78} \\
Head pose style Consistency & \revision{1.16}  & \revision{2.17} & \revision{4.70} & \revision{\textbf{3.64}} \\
\bottomrule 
\end{tabular}}
\label{table:user_study_head_pose}
\end{table*}

\subsubsection{Style Manipulation}
Thanks to the meaningful style space, our method can edit the speaking styles by manipulating style codes. As shown in Figure~\ref{fig:interpolation} and Figure~\ref{fig:interpolation_pose}, when linearly interpolating between two style codes extracted from unseen style clips, the speaking styles of generated videos transition smoothly. Through interpolation, our method is able to  control the style intensity (by interpolating the style with a neutral style) and create new speaking styles. 

\subsection{User Study}\label{user_study}

We conduct two groups of user studies involving \revision{36} participants. In the first group, we ask participants to rate the visual effects of generated videos using various methods, including Wav2Lip, PC-AVS, AVCT, EAMM, GC-AVT, ground truth, and our method. We generate three videos for each method and use the ground truth video as the expression style reference. Participants rate each video on a scale of 1-5 for lip sync quality, realness of results, and expression style consistency between the generated videos and the style reference. The mean scores are reported in Table \ref{table:user_study}. Our method outperforms existing methods in all aspects, particularly in style consistency.

In the second group, we ask participants to rate the head motions in videos generated by MakeitTalk, Audio2Head, ground truth, and our method. We also generate three videos for each method and we use the ground truth video as the head pose style reference. Participants rated each video on a scale of 1-5 for the naturalness of head motion, the synchronization between head pose and audio rhythm (head motion sync), and the consistency of head pose style between the generated videos and the style reference. The mean scores are reported in Table \ref{table:user_study_head_pose}. MakeitTalk receives low scores on all questions, indicating that it fails to produce natural-looking head movements. Although Audio2Head achieves competitive scores with our method on head motion naturalness and head motion sync, it is unable to control the style of the generated movements, thus failing to create diverse head motion sequences. These comparisons further validate the superiority of our approach.

\begin{figure}
\centering
\includegraphics[width=0.48\textwidth]{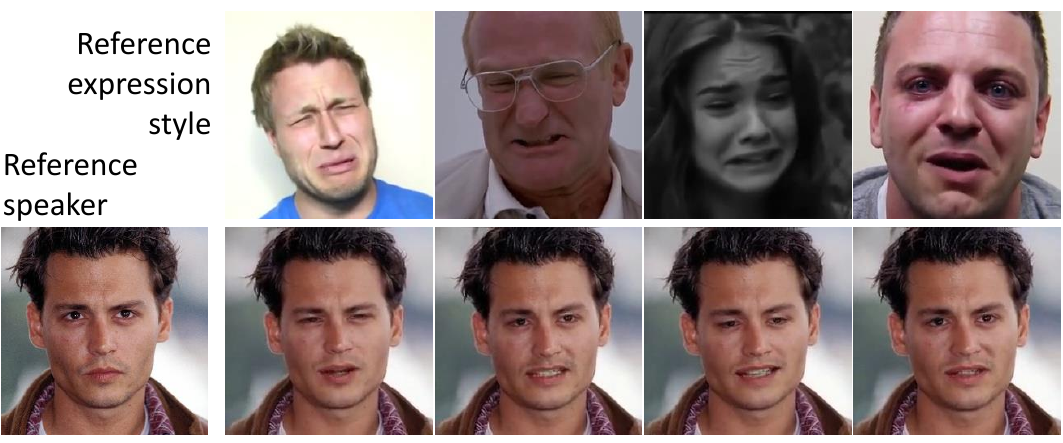}
\caption{Results with the personalized sad expression styles. The style reference videos are collected from internet sources. For more details, please zoom in or watch our demo video.
}
\label{fig:more_results}
\end{figure}

\begin{figure}
\centering
\includegraphics[width=0.48\textwidth]{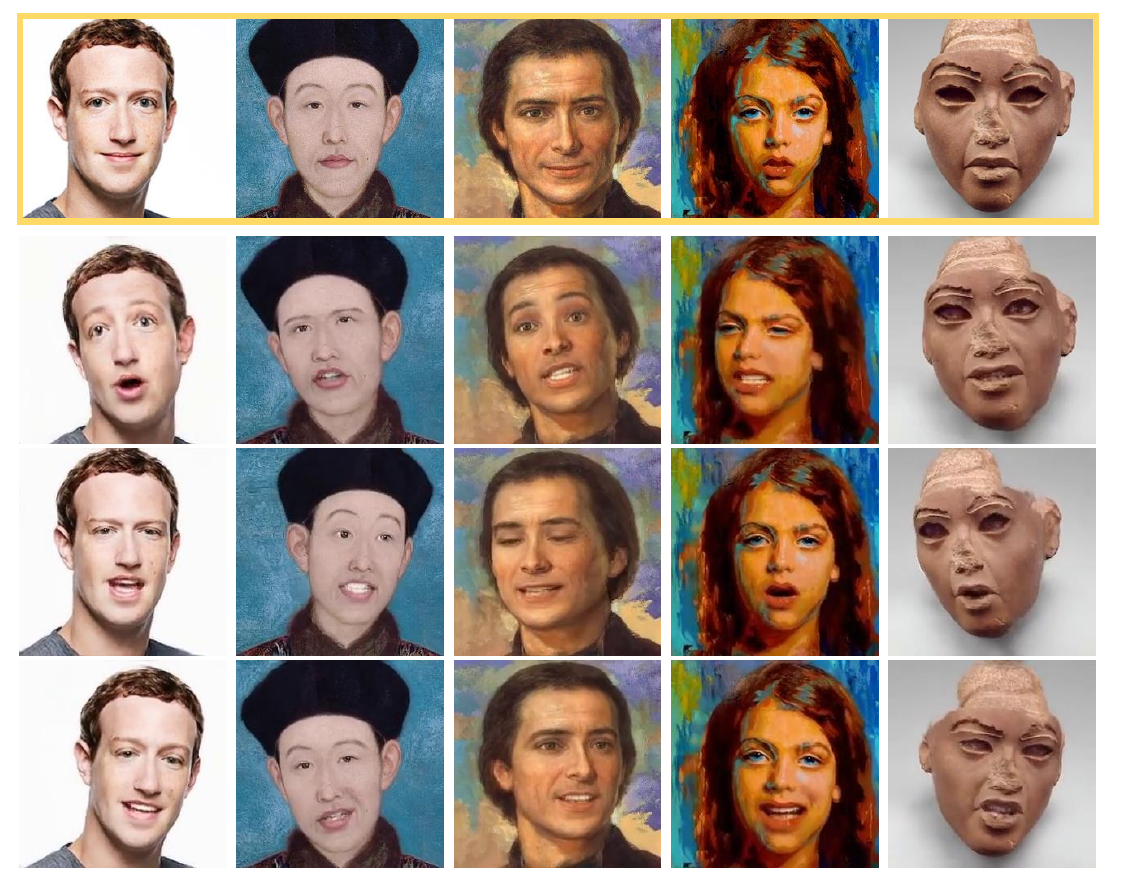}
\caption{More results generated by our method. Each video is generated with a random expression style and a random head pose style. In this figure, we display one frame from each video. The top row shows the corresponding reference speakers.
}
\label{fig:non_photo}
\end{figure}

\subsection{Discussion}\label{discussion}
\subsubsection{Further Analysis}
Figure~\ref{fig:more_results} displays the results obtained using personalized sad expression styles. The figure illustrates that our method can capture subtle differences in the motion patterns of personalized expression styles and showcase them in the generated videos. Additionally, our method successfully learns the spatio-temporal representations of previously unseen speaking styles, which is different from methods such as GC-AVT that only transfer the static expressions of reference videos. Furthermore, Figure~\ref{fig:non_photo} displays additional results produced by our method, demonstrating our method's capability of animating talking head videos for non-photorealistic paintings. This showcases the promising generation ability of our method beyond face photography.


Our method has the capability to generate expressive talking videos in real-time by utilizing just an audio clip, a reference speaker image, and alternative style reference video clips. This technology has the ability to generalize to unseen style clips and various types of facial photography. This feature opens up many interesting practical applications, including the creation of short videos and visual dubbing. This technology can be particularly useful in the entertainment industry, where it can be used to dub a foreign language film or TV show into another language while maintaining the same emotional intensity and facial expressions. Additionally, this technology can be used to create personalized videos for individuals, such as a special greeting from a celebrity or a personalized message from a loved one.


\subsubsection{Limitation and Future Work}
While our method produces convincing results, there are still some limitations. \revision{Firstly, our method fails to extract reasonable expression styles from style reference videos with extreme head poses and side views. Additionally, in extreme expressions, our method doesn't always fully close the lips for some phonemes, such as \textit{p}, \textit{b}, and \textit{m}. Secondly, our approach is also restricted by the length of the reference video. We find that when the length of the expression reference video is less than 0.5 seconds, or the length of the head pose reference video is less than 3 seconds, the network has difficulty extracting the appropriate style. Furthermore, if there is a significant style difference within the same reference video, our method yields uncertain results. }

Since we completely remove the articulation-irrelevant information in the stylized expression generation branch, including the audio rhythm, there is a possibility that the facial expression rhythm in the synthetic video does not match the audio. In our future work, we plan to disentangle the audio rhythm information and integrate it into our framework.

Furthermore, limited by the image renderer, our methods may produce artifacts around the mouth when generating intense facial expressions, and around the head when generating large head movements. In our future work, we aim to expand the emotional talking video datasets and develop more advanced rendering techniques.

\section{Conclusion}

In this paper, we introduce a new framework called \textit{StyleTalk++}, which generates one-shot audio-driven talking faces with diverse speaking styles. Our method effectively extracts expression styles and head pose styles from arbitrary style reference videos, and then injects them into the generated talking face videos using a unified style-controllable framework. In contrast to previous works, our approach captures the spatio-temporal co-activations of speaking styles from the style reference videos, leading to authentic stylized talking face videos. Extensive experiments show that our method produces photo-realistic talking head videos with conditional speaking styles while achieving more accurate lip-sync and better identity preservation than the state-of-the-art.

\ifCLASSOPTIONcompsoc
  \section*{Acknowledgments}
\else
  \section*{Acknowledgment}
\fi

This work is supported by the 2022 Hangzhou Key Science and Technology Innovation Program (No. 2022AIZD0054) and the Key Research and Development Program of Zhejiang Province (No. 2022C01011). This research is partially funded by the ARC-Discovery grants (DP220100800) and ARC-DECRA (DE230100477). This work was supported in part by the National Science Foundation of China (NSFC) under Grant No. 62176134, by a grant from the Institute Guo Qiang (2019GQG0002), Tsinghua University, and by research and application on AI technologies for smart mobility funded by SAIC Motor.

We would like to express our gratitude to Xinya Ji, Borong Liang, and Yan Pan for generously assisting us with the comparisons. We also thank Ran Yi, Zipeng Ye, and Zhiyao Sun for sharing their HeadMotion dataset with us. Additionally, we would like to thank Lincheng Li and Zhimeng Zhang for their valuable contributions to our discussions.

\ifCLASSOPTIONcaptionsoff
  \newpage
\fi



\bibliographystyle{IEEEtran}
%


\bibliography{tpami}

%

\begin{IEEEbiography}[{\includegraphics[width=1in,height=1.25in,clip,keepaspectratio]{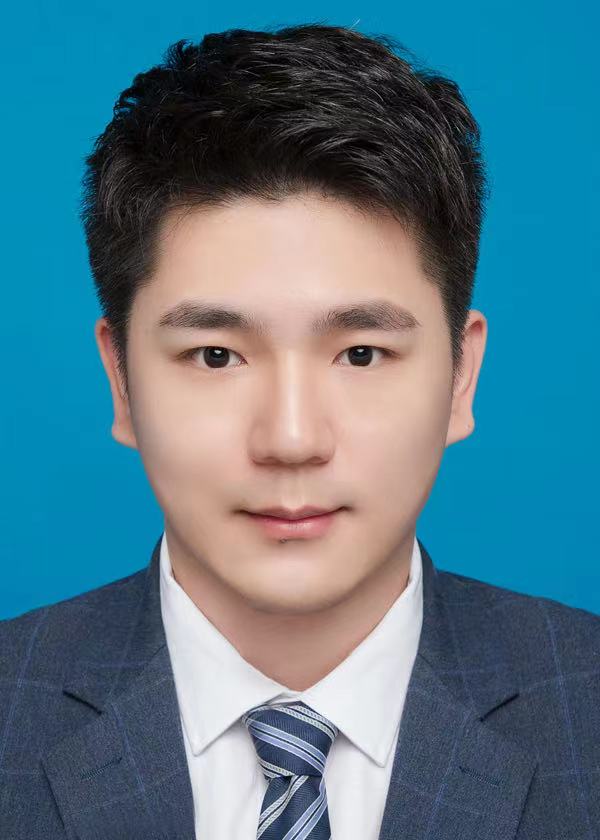}}]{Suzhen Wang} is currently an artificial intelligence researcher at Netease Fuxi AI Lab, Hangzhou, China. He received B.E. degree from Southeast University, Nanjing, China, in 2016, and the M.S. degree from Zhejiang University, Hangzhou, China in 2019. His research interests include computer vision, multimodal learning, image and video processing, animation generation and their application in games.
\end{IEEEbiography}

\begin{IEEEbiography}[{\includegraphics[width=1in,height=1.25in,clip,keepaspectratio]{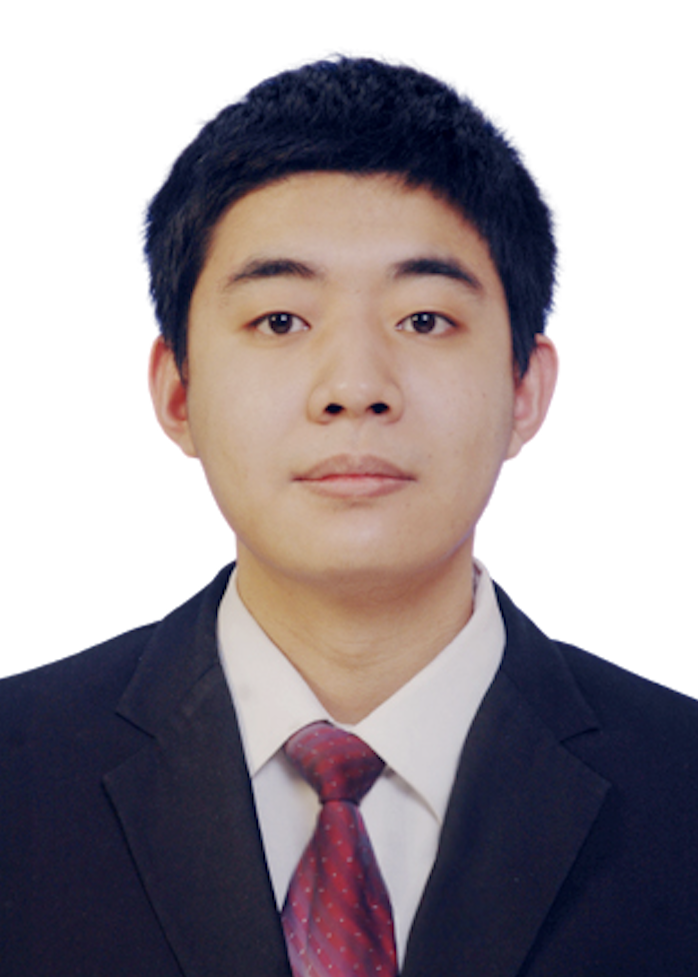}}]{Yifeng Ma}
is a PhD student with Department
of Computer Science and Technology, Tsinghua
University. He received his B.Eng. degree from
Harbin Institute of Technology, China, in 2018. His research interest focuses on computer vision and machine
intelligence.
\end{IEEEbiography}
\begin{IEEEbiography}[{\includegraphics[width=1in,height=1.25in,clip,keepaspectratio]{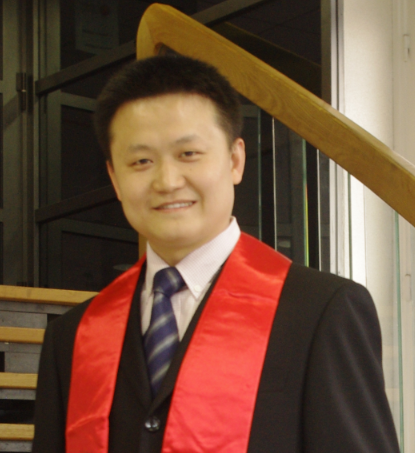}}]{Yu Ding} is currently an artificial intelligence expert, leading the virtual human team at Netease Fuxi AI Lab, Hangzhou, China. His research interests include virtual human, deep learning, image and video processing, talking-head generation, animation generation, multimodal computing, affective computing, and embodied conversational agent. He received Ph.D. degree in Computer Science at Telecom Paristech in Paris (France). 
\end{IEEEbiography}

\begin{IEEEbiography}[{\includegraphics[width=1in,height=1.25in,clip,keepaspectratio]{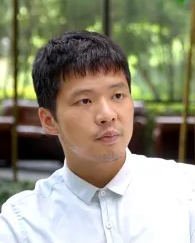}}]{Zhipeng Hu} - Vice president of NetEase Group. He graduated from Zhejiang University and once worked at MSRA. His major interest includes Game Design, Game AI, Computer Vision and Graphics. He is also the co-founder of NetEase Fuxi AI Lab and has great insight into the cross-domain of game and AI. He has be responsible for developing several most successful and popular games in worldwide, such as Ghost and Justice. \end{IEEEbiography}
\vspace{-10 mm} 
\begin{IEEEbiography}[{\includegraphics[width=1in,height=1.25in,clip,keepaspectratio]{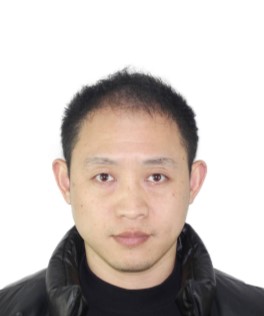}}]{Changjie Fan}
    is the Director of NetEase Fuxi AI Lab. He received his doctor’s degree in Computer Science from University of Science and Technology of China. His research interest is in machine learning, including multiagent systems, deep reinforcement learning, game theory and knowledge discovery.
\end{IEEEbiography}
\begin{IEEEbiography}[{\includegraphics[width=1in,height=1.25in,clip,keepaspectratio]{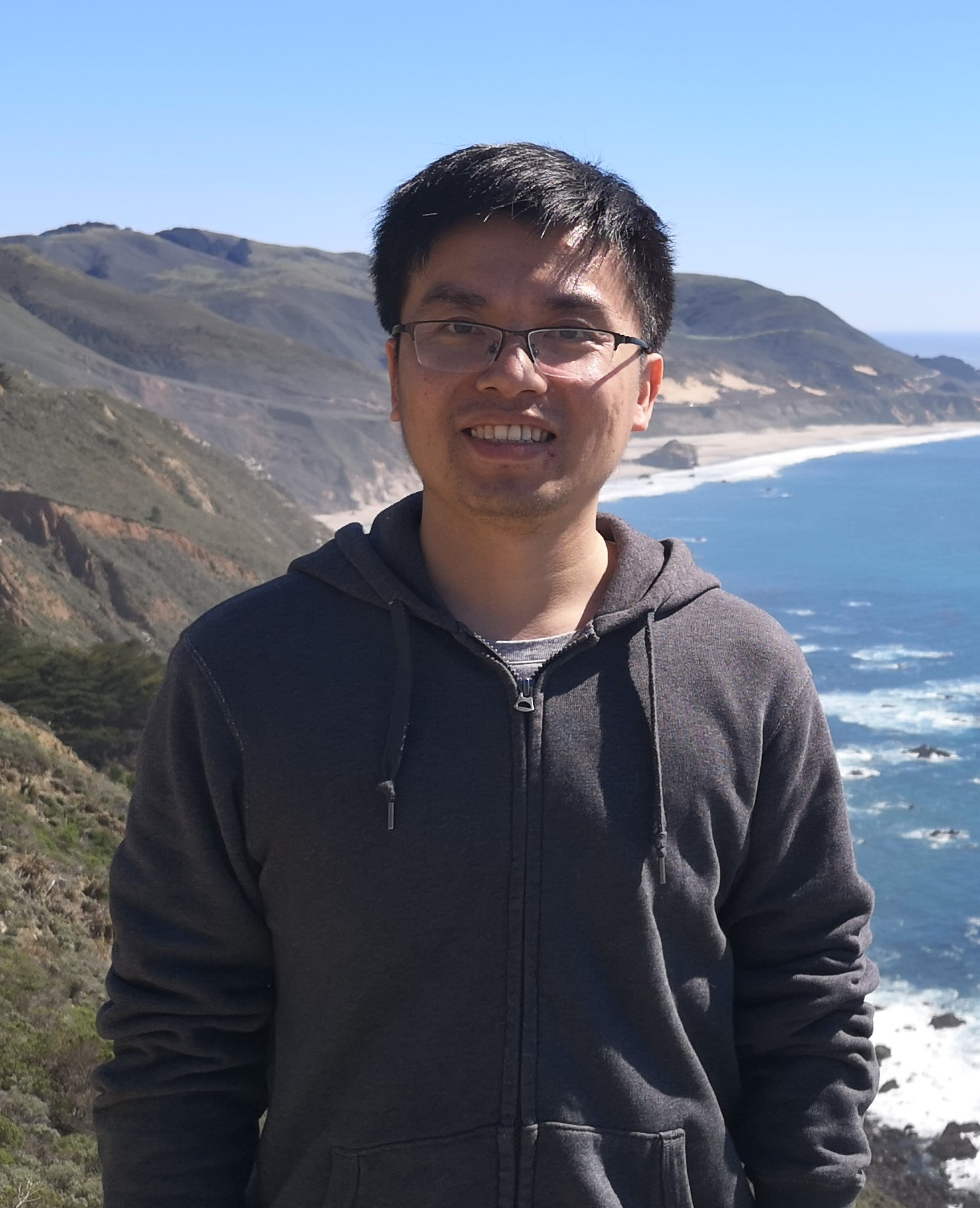}}]{Tangjie Lv} is the head of the AI team of NetEase Fuxi Lab. He received his PhD in computational mathematics from Peking University. After graduation, he worked for NetEase Games and Tencent AI Lab. At present, he is responsible for the AI research and technology development in games and other entertainment scenes. His research interests include reinforcement learning, machine learning and game ai. He has led his team to achieve successful AI implementation in several NetEase games.
\end{IEEEbiography}
\begin{IEEEbiography}[{\includegraphics[width=1in,height=1.25in,clip,keepaspectratio]{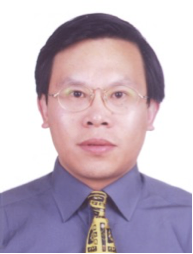}}]{Zhidong Deng}
is a professor with the Department of Computer Science and Technology, Tsinghua University, Beijing, China. He received the B.S. degree from Sichuan University, Chengdu, China, in 1986 and the Ph.D. degree from Harbin Institute of Technology, Harbin, China, in 1991, respectively, both in computer science and automation. He has been a Full Professor at Tsinghua University since 2000. His current research areas include artificial intelligence, computational neuroscience, autonomous driving, and advanced robotics. 
\end{IEEEbiography}

\begin{IEEEbiography}[{\includegraphics[width=1in,height=1.25in,clip,keepaspectratio]{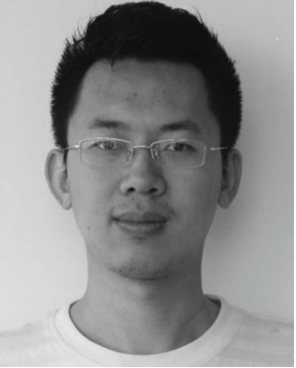}}]{Xin Yu} received the B.S. degree in electronic engineering from the University of Electronic Science and Technology of China, Chengdu, China, in 2009, the Ph.D. degree from the Department of Electronic Engineering, Tsinghua University, Beijing, China, in 2015, and the Ph.D. degree from the College of Engineering and Computer Science, Australian National University, Canberra, Australia, in 2019. He is currently a Senior Lecturer at the University of Queensland. His research interests include computer vision and image processing. 
\end{IEEEbiography}

\end{document}